\documentclass[lettersize,journal]{IEEEtran}
\usepackage{amsmath,amsfonts}
\usepackage{algorithmic}
\usepackage{algorithm}
\usepackage{array}
\usepackage[caption=false,font=normalsize,labelfont=sf,textfont=sf]{subfig}
\usepackage{textcomp}
\usepackage{stfloats}
\usepackage{threeparttable}
\usepackage{url}
\usepackage{enumerate}
\usepackage{verbatim}
\usepackage{graphicx}
\usepackage{cite}
\usepackage{amsmath}
\usepackage{amsmath, bm}
\usepackage{bbding}
\usepackage{pifont}
\usepackage{booktabs}

\usepackage{amsfonts}
\usepackage{hyperref}
\hypersetup{
    colorlinks=true,            
    linkcolor=blue,              
    citecolor=blue,            
    filecolor=mycustompurple,   
    urlcolor=magenta            
}

\usepackage{multirow}
\usepackage{dsfont}
\usepackage{color}
\usepackage[table]{xcolor}
\usepackage{bbm}
\usepackage{textcomp}
\begin{document}
\title{ViRefSAM: Visual Reference-Guided Segment Anything Model for Remote Sensing Segmentation}

\author{
 Hanbo~Bi,
 Yulong~Xu,
 Ya~Li,
 Yongqiang~Mao,
 Boyuan~Tong,
 Chongyang~Li,
 Chunbo~Lang, 
 Wenhui~Diao,
 Hongqi~Wang,~\IEEEmembership{Member,~IEEE,}
 Yingchao~Feng,~\IEEEmembership{Member,~IEEE,}
 and~Xian~Sun,~\IEEEmembership{Senior Member,~IEEE}
\thanks{This work for Remote Sensing Segmentation was supported by the National Natural Science Foundation of China (NSFC) under Grant 62301538. (\emph{Corresponding author: Yingchao Feng.})} 

\thanks{H. Bi, Y. Xu, B. Tong, C. Li, and X. Sun are with the Aerospace Information Research Institute, Chinese Academy of Sciences, Beijing 100190, China, also with the School of Electronic, Electrical and Communication Engineering, University of Chinese Academy of Sciences, Beijing 100190, China, also with the University of Chinese Academy of Sciences, Beijing 100190, China, and also with the Key Laboratory of Target Cognition and Application Technology(TCAT), Aerospace Information Research Institute, Chinese Academy of Sciences, Beijing 100094, China.}
 
\thanks{Y. Li, W. Diao, H. Wang, and Y. Feng are with the Aerospace Information Research Institute, Chinese Academy of Sciences, Beijing 100190, China, and also with the Key Laboratory of Target Cognition and Application Technology(TCAT), Aerospace Information Research Institute, Chinese Academy of Sciences, Beijing 100094, China.}

\thanks{Y. Mao is with the Department of Electronic Engineering, Tsinghua University, Beijing 100084, China.}
\thanks{C. Lang is with the School of Automation, Northwestern Polytechnical University, Xi’an 710129, China.}

}

\markboth{Journal of \LaTeX\ Class Files,~Vol.~14, No.~8, August~2021}%
{Shell \MakeLowercase{\textit{et al.}}: A Sample Article Using IEEEtran.cls for IEEE Journals}


\maketitle

\begin{abstract}\label{abstract}
The Segment Anything Model (SAM), with its prompt-driven paradigm, exhibits strong generalization in generic segmentation tasks. However, applying SAM to remote sensing (RS) images still faces two major challenges. First, manually constructing precise prompts for each image (e.g., points or boxes) is labor-intensive and inefficient, especially in RS scenarios with dense small objects or spatially fragmented distributions. Second, SAM lacks domain adaptability, as it is pre-trained primarily on natural images and struggles to capture RS-specific semantics and spatial characteristics, especially when segmenting novel or unseen classes. To address these issues, inspired by few-shot learning, we propose ViRefSAM, a novel framework that guides SAM utilizing only a few annotated reference images that contain class-specific objects. Without requiring manual prompts, ViRefSAM enables automatic segmentation of class-consistent objects across RS images. Specifically, ViRefSAM introduces two key components while keeping SAM’s original architecture intact: (1) a Visual Contextual Prompt Encoder that extracts class-specific semantic clues from reference images and generates object-aware prompts via contextual interaction with target images; and (2) a Dynamic Target Alignment Adapter, integrated into SAM’s image encoder, which mitigates the domain gap by injecting class-specific semantics into target image features, enabling SAM to dynamically focus on task-relevant regions. Extensive experiments on three few-shot segmentation benchmarks, including iSAID-5$^i$, LoveDA-2$^i$, and COCO-20$^i$, demonstrate that ViRefSAM enables accurate and automatic segmentation of unseen classes by leveraging only a few reference images, and consistently outperforms existing few-shot segmentation methods across diverse datasets.

\end{abstract}


\section{Introduction}\label{Indroduction}
\IEEEPARstart{S}{emantic} segmentation is a fundamental task in remote sensing (RS) intelligent interpretation systems~\cite{csillik2017fast,kotaridis2021remote,diakogiannis2020resunet}, aimed at assigning pre-defined semantic labels to each pixel in RS images for accurate identification and spatial localization of land cover categories. This task plays a critical role in various real-world applications, including land use monitoring~\cite{boonpook2023deep,toker2022dynamicearthnet}, urban planning~\cite{weng2012remote,pan2020deep}, and disaster assessment~\cite{zhu2021msnet,asad2023natural}. With the rapid development of deep learning, particularly with the emergence of fully convolutional networks (FCNs)~\cite{long2015fully,mohammadimanesh2019new,pastorino2022semantic} and Transformer-based architectures~\cite{dosovitskiy2020image,wang2022unetformer,cheng2022masked}, the accuracy and generalization capabilities of RS semantic segmentation have substantially improved, providing a solid foundation for the automated analysis of RS images.

\begin{figure}[t]
\setlength{\abovecaptionskip}{1pt}
\centering
\includegraphics[width=1.0\linewidth]{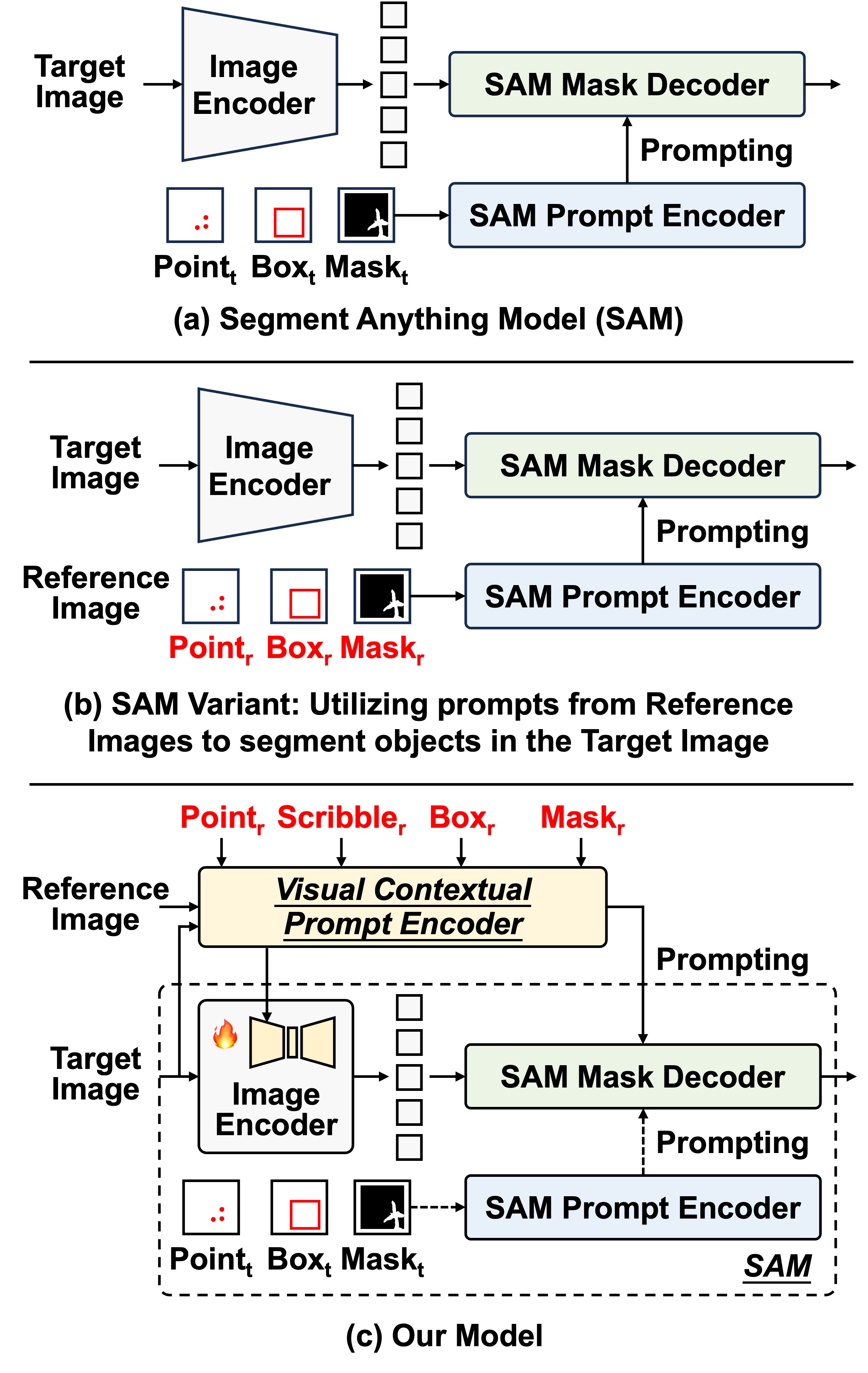}
\caption{ \textbf{Framework comparison among SAM, SAM variant, and our ViRefSAM.} (a) \textbf{Original SAM:} Requires manually customized prompts (e.g., points, boxes, or masks) for each individual image, which must be carefully constructed to activate the correct target. (b) \textbf{SAM Variant:} Attempts to reduce interaction cost by reusing prompts (e.g., a box or point) defined in one reference image across other images. However, this scheme lacks semantic alignment between images and may fail under significant variations in object appearance and background. (c) \textbf{Our ViRefSAM:} Introduces a Visual Contextual Prompt (VCP) Encoder to automatically generate object-aware prompts by encoding high-level semantic information from the reference image and dynamically interacting with the target image features. This design eliminates the need for manually customized or reused prompts while preserving SAM’s core architecture.}
\label{fig:1}
\end{figure}

The Segment Anything Model (SAM) has recently emerged as a highly influential vision foundation model for generic segmentation~\cite{kirillov2023segment,zhao2023fast,chen2023sam,wu2025medical}. Pre-trained on billions of labeled samples, SAM demonstrates impressive zero-shot generalization across diverse segmentation tasks. Its core innovation is the prompt-driven paradigm, which allows users to specify target regions flexibly using point, bounding box, or mask prompts for efficient and accurate segmentation (see Fig.\ref{fig:1}(a)). Motivated by SAM’s capabilities, researchers have begun applying it to RS semantic segmentation~\cite{wang2023samrs,chen2024rsprompter,ringmosam,ma2024sam,qiao2025sam}. For example, Wang et al.~\cite{wang2023samrs} proposed an efficient pipeline that combines SAM with existing RS object detection datasets to generate large-scale RS segmentation datasets. Similarly, RingMo-SAM~\cite{ringmosam} proposed a multi-modal RS segmentation foundation model that segments objects in optical and SAR data while also identifying object categories. SSRS~\cite{ma2024sam} introduced object consistency and boundary preservation losses to refine SAM’s output for segmentation tasks.

\begin{figure}[t]
\setlength{\abovecaptionskip}{1pt}
\centering
\includegraphics[width=1.0\linewidth]{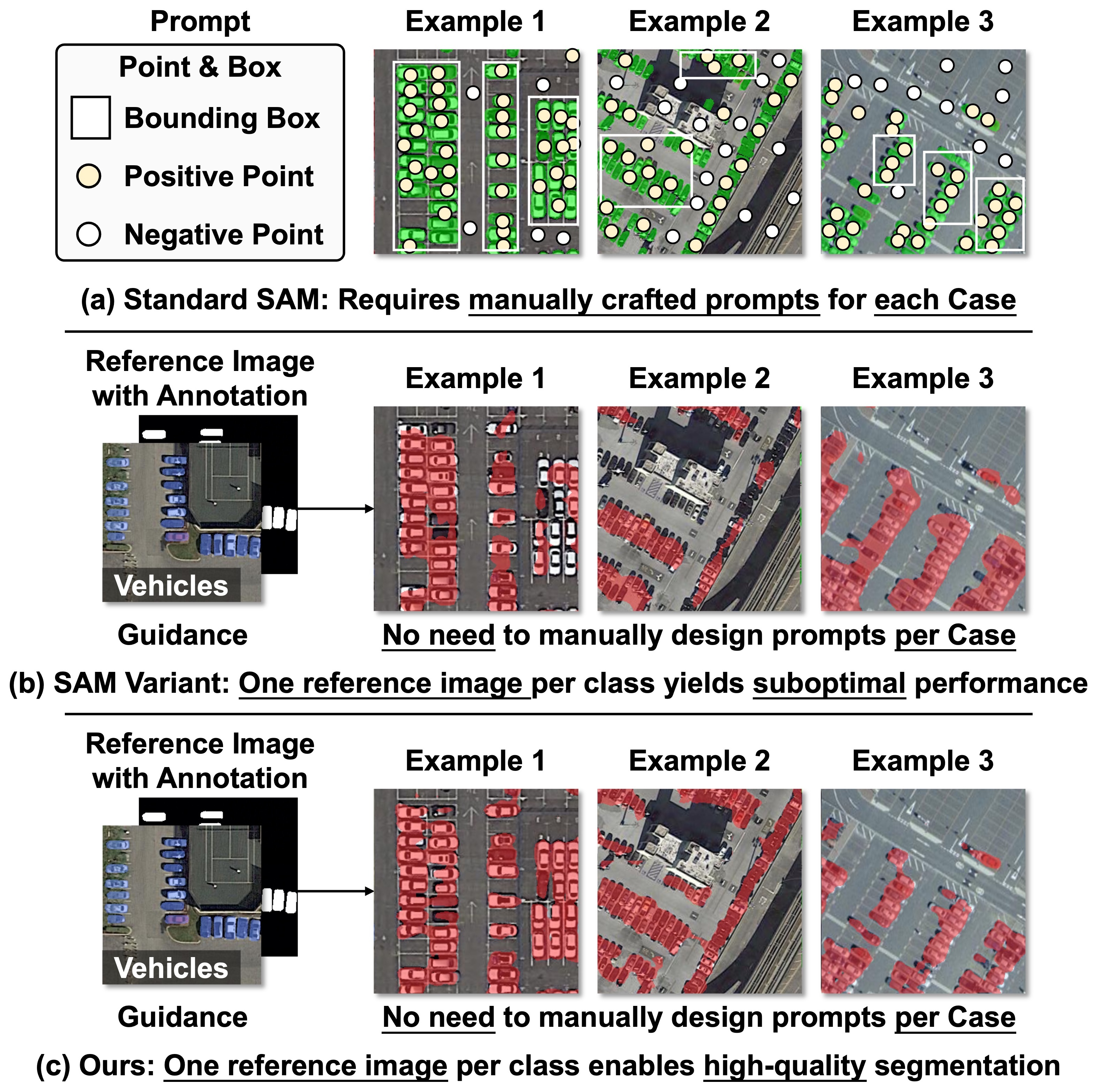}
\caption{ \textbf{Comparison of segmentation workflow among SAM, SAM variant, and our ViRefSAM.} (a) \textbf{SAM:} Requires manually crafted prompts (e.g., points or boxes) for each image, making it inefficient in large-scale RS segmentation. (b) \textbf{SAM variant:} Reuses prompts generated from a single labeled reference image per class, but lacks task-specific alignment, resulting in limited segmentation quality. (c) \textbf{Our ViRefSAM:} Introduces a Visual Contextual Prompt Encoder and Dynamic Target Alignment Adapter to fully exploit annotated reference images, enabling accurate and automatic segmentation of specific classes across RS images.
 }
\label{fig:2}
\end{figure}

Despite its promising performance, the application of SAM to remote sensing (RS) imagery still faces several key challenges. \textbf{1) Prompt Construction Efficiency.} SAM requires manually constructing precise and reasonable prompts (e.g., points, boxes, or rough masks) for each RS image, which is both time-consuming and labor-intensive (see Fig.\ref{fig:2}(a)). To segment specific objects, users must first accurately localize targets within complex scenes and then provide the corresponding prompts to activate SAM. The more precise and reasonable these prompts are, the more effective the segmentation. This process becomes particularly cumbersome in RS scenarios involving challenging spatial structures and dense small objects, such as “vehicles” densely packed in parking lots, “ships” scattered across open waters, or “buildings” in urban residential areas. These targets often exhibit high intra-class variability, tiny object scale, and cluttered or fragmented spatial distributions, significantly increasing the manual annotation burden. \textbf{2) Domain-specific Knowledge.} Since SAM is predominantly pre-trained on natural scene images, it lacks the domain-specific semantic and spatial understanding required for RS imagery, leading to degraded segmentation performance. This limitation becomes more pronounced when encountering novel or unseen RS-specific classes not present in the pre-training corpus, restricting its generalization in real-world applications.

To enhance both efficiency and generalization of SAM in RS segmentation, the few-shot learning paradigm~\cite{vinyals2016matching,bi2023not,sung2018learning,wang2020generalizing} offers a promising solution. This paradigm enables models to learn novel patterns from only a few labeled examples. In this context, visual contextual samples (i.e., reference images) can serve as semantic prompts to guide SAM in segmenting desired objects. As illustrated in Fig.\ref{fig:2}(b)-(c), by providing a single reference image containing instances of “vehicles”, SAM can automatically segment “vehicles” in various RS images without requiring manually crafted prompts for each image. A straightforward engineering implementation is to reuse prompts across images. As illustrated in Fig.\ref{fig:1}(b), point or box annotations on a reference image can be encoded by SAM's Prompt Encoder and directly applied to segment “vehicles” in other images via SAM's Mask Decoder. However, such static prompt reuse often overlooks the semantic gap between RS images caused by significant differences in object appearance, scale, and background complexity, which ultimately hinders generalization performance.

To address these limitations, we propose ViRefSAM, a visual reference-guided segmentation framework that leverages visual reference images of a specific category to automatically segment class-consistent objects across diverse RS images. Unlike previous schemes that rely on manually customized or statically reused prompts, ViRefSAM introduces a Visual Contextual Prompt (VCP) Encoder that learns high-level, class-specific semantic representations from the reference image and dynamically interacts with the target image features. This enables the generation of object-aware prompts in a context-sensitive manner, without modifying SAM’s core architecture. As illustrated in Fig.\ref{fig:1}(c), the reference image and its corresponding annotations (supporting various formats such as points, scribbles, bounding boxes, and dense masks) are fed into the VCP Encoder to generate semantic prompts, which are then forwarded to SAM’s Mask Decoder to activate the desired target objects.

The VCP encoder operates in two primary stages: visual contextual interaction and object-aware prompt generation. (a) Visual contextual interaction: Initially, the reference image is condensed into a set of reference prototypes guided by the provided annotations, which effectively capture semantic clues of class-specific objects. These prototypes then engage with the target image features (the image to be segmented) via a cross-attention structure, enabling the extraction of class-specific semantic information directly from the target image. The target image’s pseudo-mask is subsequently derived by aligning the reference prototypes with the target image features, ensuring accurate object localization. (b) Object-aware prompt generation: A set of learnable queries is introduced to perform mask attention, constrained by the generated pseudo-mask, enabling the extraction of object-specific semantics from the target image. These representations culminate in the generation of object-aware embeddings that facilitate segmentation. To enhance the generation quality and prevent semantic redundancy, we incorporate a regularization loss, ensuring that the embeddings are distinct and contextually relevant.

To further enhance domain adaptation for RS target segmentation, we fine-tune SAM within a meta-learning framework~\cite{finn2017model}. The training set is organized into multiple tasks (i.e., episodes), each focusing on a specific object category. This forces the model to learn class-specific contextual information from reference images to segment corresponding objects in target images. For this purpose, we propose a Dynamic Target Alignment (DTA) Adapter. Unlike traditional adapters~\cite{sung2022vl} that adjust image features through downsampling and upsampling projections, the DTA adapter introduces a parameter-free unit that dynamically embeds reference image prototypes, containing class-specific semantic clues, into the target image features, under the constraint of positional encoding. This injection encourages SAM to activate class-specific objects relevant to the current task. With this fine-tuning strategy, SAM can quickly and accurately focus on novel category objects in RS scenes by leveraging reference images, without compromising its generalization ability.

For evaluation, we followed the standard few-shot segmentation (FSS) protocol~\cite{wang2019panet,liu2020part,lang2024few}, which utilizes several reference images to guide the model in segmenting unseen classes. We conducted experiments on two RS FSS datasets, iSAID-5$^i$ and LoveDA-2$^i$, as well as the CV dataset COCO-20$^i$. Results show that ViRefSAM effectively mitigates the limitations of manual prompt engineering in SAM, achieving automatic and accurate segmentation of class-specific objects, even for unseen classes, guided by only a few reference images. It outperforms existing FSS methods, including vision foundation models such as PerSAM and Matcher, across all three datasets, delivering the best results overall.

The main contributions of our work can be summarized as follows:

\begin{enumerate}
\item We enhance the functionality of SAM by proposing the ViRefSAM framework, which leverages visual reference images of a specific category to automatically segment the class-consistent objects across diverse images, eliminating the need for manually customized prompts for each image.

\item We introduce a Visual Contextual Prompt (VCP) Encoder that extracts class-specific semantic representations from reference images and dynamically interacts with target image features to generate object-aware prompts, which are seamlessly integrated into SAM’s mask decoder for segmentation.

\item We propose a Dynamic Task Alignment Adapter, which, in conjunction with the meta-learning training paradigm, drives the SAM model to dynamically focus on class-specific objects within the current task, enhancing the generalization capability in RS scenarios.
\item Extensive experiments on RS and CV few-shot segmentation datasets demonstrate the effectiveness of ViRefSAM. With only a few reference images, it enables automatic segmentation of class-specific objects (even for unseen classes) and outperforms existing few-shot segmentation methods across multiple datasets.

\end{enumerate}

\section{Related work}\label{Related work}
\subsection{Remote Sensing Semantic Segmentation}
Semantic segmentation, which assigns a semantic label to each pixel, is a fundamental task in RS image interpretation. Long et al.~\cite{long2015fully} pioneered Fully Convolutional Networks (FCNs) for dense prediction, laying the groundwork for subsequent developments. To better capture contextual information, Chen et al.~\cite{chen2014semantic} introduced dilated convolutions to expand the receptive field without resolution loss, while modules like the Pyramid Pooling Module (PPM)~\cite{zhao2017pyramid} and Atrous Spatial Pyramid Pooling (ASPP)~\cite{chen2017deeplab} enabled effective multi-scale feature fusion. In the RS domain, the complexity of scenes, the prevalence of small targets, and class imbalance pose additional challenges. To address these, Peng et al.~\cite{peng2021cross} proposed CF-Net with multi-branch fusion to enhance small-object representation. Feng et al.~\cite{feng2020npaloss} introduced NPALoss to improve boundary delineation and object localization. Niu et al.~\cite{9580861} further proposed a disentangled framework to separately model foreground regions and object boundaries. While these CNN-based methods achieve strong performance on large-scale datasets, their generalization to unseen categories or domains remains limited, especially under low-data conditions.

Recently, Transformer-based architectures have advanced semantic segmentation by modeling long-range dependencies and global context. SegFormer~\cite{xie2021segformer} and Mask2Former~\cite{cheng2022masked} have demonstrated strong scalability in natural scenes. Inspired by these advances, RS studies have introduced tailored Transformer designs~\cite{he2022swin,wang2022unetformer,xiao2023enhancing,liu2023rethinking}. For instance, UNetFormer~\cite{wang2022unetformer} employs a Transformer-based decoder for real-time urban segmentation, and Xiao et al.~\cite{xiao2023enhancing} proposed a hybrid CNN-Transformer architecture to enhance multiscale feature learning.  However, these methods still rely on extensive annotations, limiting their generalization in few-shot and cross-domain tasks.

\subsection{Segment Anything Model}
The Segment Anything Model (SAM)~\cite{kirillov2023segment}, proposed by Meta AI, is a universal interactive segmentation framework built upon the vision transformer architecture. Pre-trained on billions of labeled samples, SAM exhibits strong zero-shot generalization capabilities across a wide range of segmentation tasks. The model comprises three core components: an image encoder, a prompt encoder, and a mask decoder. Specifically, the image encoder extracts high-level visual features from the input image; the prompt encoder converts user-specified inputs (e.g., points, boxes, or masks) into prompt embeddings; and the mask decoder integrates these embeddings to generate accurate segmentation masks. Due to its versatility, SAM has been rapidly adopted across various domains beyond traditional semantic segmentation. In the medical imaging domain, MedSAM~\cite{ma2024segment} constructs a large-scale dataset covering 10 imaging modalities and over 30 cancer types to support the development of a universal medical segmentation model. In image editing and restoration, frameworks such as Inpaint Anything~\cite{yu2023inpaint} and Edit Everything~\cite{xie2023edit} employ SAM-generated masks to define editable regions with high spatial precision. Additionally, SAM has been integrated into object tracking~\cite{cheng2023segment,rajivc2025segment}, where prompt-guided initialization improves robustness in target localization. These applications highlight SAM’s potential as a vision foundation model, particularly in data-scarce or weakly supervised settings.

\begin{figure*}[t]
\setlength{\abovecaptionskip}{1pt}
\centering
\includegraphics[width=1.0\linewidth]{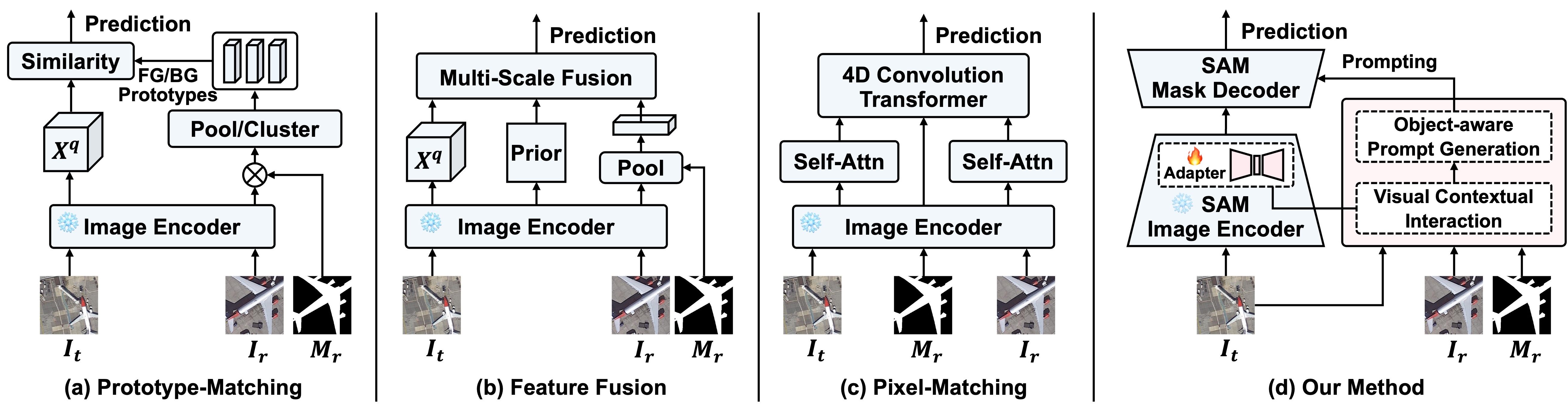}
\caption{\textbf{Overview of typical Few-shot Segmentation (FSS) paradigms.} (a) \textbf{Prototype matching-based methods}~\cite{zhang2020sg, liu2020part, fan2022self}, compute class prototypes from reference images and compare them with target features. (b) \textbf{Feature fusion-based methods}~\cite{min2021hypercorrelation, cheng2022holistic, liu2022learning}, aggregate reference and target features to enhance class-specific object representation. (c) \textbf{Pixel matching-based methods}~\cite{zhang2021few, wang2022adaptive, xu2023self, peng2023hierarchical}, establish pixel-wise correlations between reference and target images for fine-grained segmentation. (d) \textbf{Our ViRefSAM}, introduces a Visual Contextual Prompt Encoder into SAM, leveraging reference images to generate object-aware prompts and enabling automatic segmentation of class-specific objects in RS imagery. This method effectively overcomes the manual prompt design burden in SAM-based RS segmentation tasks.}
\label{fig:3}
\end{figure*}

\subsection{Segment Anything Model in Remote Sensing}
In the RS domain, SAM has garnered increasing attention for its potential to reduce annotation costs and enhance segmentation quality in large-scale earth observation tasks. Recent efforts fall into two main directions. The first focuses on adapting SAM with RS-specific knowledge to improve performance in RS imagery. For instance, MeSAM~\cite{MeSAM} introduces a multi-scale enhancement strategy to address SAM’s limitations in boundary and texture representation. RingMo-SAM~\cite{ringmosam} develops a multimodal segmentation foundation model for optical and SAR data, and RSPrompter~\cite{chen2024rsprompter} incorporates semantic category priors into SAM for automated instance segmentation. The second direction leverages SAM as a source of high-quality priors to support downstream tasks, such as road extraction and building delineation~\cite{luo2024sam,feng2024road,wu2025tpp}. Qi et al.~\cite{qi2024multi} inject SAM-derived features into implicit neural fields for multi-view segmentation under limited supervision. SAM-CD~\cite{ding2024adapting} applies SAM to change detection via prompt-based segmentation, and Wang et al.~\cite{wang2023samrs} propose a scalable pipeline that combines SAM with detection datasets to automatically generate segmentation annotations. 

\subsection{Few-shot (Visual Reference) Segmentation}\label{Few-shot segmentation}
Recent advancements in few-shot learning (FSL) have aimed at improving generalization to unseen domains, enabling models to effectively classify novel classes with only a few labeled samples. The majority of current methods adopt a meta-learning paradigm~\cite{finn2017model}, where transferable knowledge is learned from a series of tasks (episodes) drawn from the training dataset and generalized to novel tasks. Few-shot segmentation (FSS) extends FSL to dense prediction tasks, aiming to segment unseen classes in target images using a small number of labeled samples (i.e., reference images)~\cite{shaban2017one}.

FSS approaches can be grouped into three primary branches: (1) Prototype matching-based methods (see Fig.\ref{fig:3}(a))~\cite{zhang2020sg,wang2019panet,liu2020part,fan2022self,lang2024few}, which generate category-representative prototypes by compressing reference features through Mask Average Pooling~\cite{zhang2020sg}. For example, SSPNet~\cite{fan2022self} aggregates prototypes from the semantic clues within the target image to guide its segmentation. DCPNet~\cite{lang2024few} refines reference features into multiple prototypes, each representing distinct properties. (2) Feature fusion-based methods (see Fig.\ref{fig:3}(b))~\cite{min2021hypercorrelation,cheng2022holistic,liu2022learning,yang2023mianet}, which combine features from both the reference and target images. For instance, HSNet~\cite{min2021hypercorrelation} utilizes 4D convolution for multi-scale fusion, while IPMT~\cite{liu2022intermediate} merges reference- and target-image semantics into an intermediate prototype to bridge discrepancies between the images. (3) Pixel matching-based methods (see Fig.\ref{fig:3}(c))~\cite{zhang2021few,wang2022adaptive,xu2023self,peng2023hierarchical}, which focus on pixel-level similarity between reference and target images. CyCTR~\cite{zhang2021few} uses cross-attention to aggregate pixel-level semantics from the reference image. 

In the RS domain, few-shot segmentation has attracted substantial attention~\cite{yao2021scale,wang2021dmml,jiang2022few,lang2023global,10152484,bi2023not}. Yao et al.~\cite{yao2021scale} proposed a scale-aware prototype matching approach to tackle the variability in object appearance and scale. Bi et al.~\cite{bi2023not} developed DMNet to extract class-specific semantics from target images. Peng et al.~\cite{MGANet} introduced a multi-granularity aggregation network to progressively capture discriminative information at multiple levels, addressing the challenges of intra-class inconsistency in RS FSS tasks. Inspired by these advances, we introduce a Visual Contextual Prompt Encoder into SAM, enabling automatic segmentation of RS objects guided by reference images and mitigating the need for costly manual prompt design.

\begin{figure*}[t]
\setlength{\abovecaptionskip}{2pt}
\centering
\includegraphics[width=1.0\linewidth]{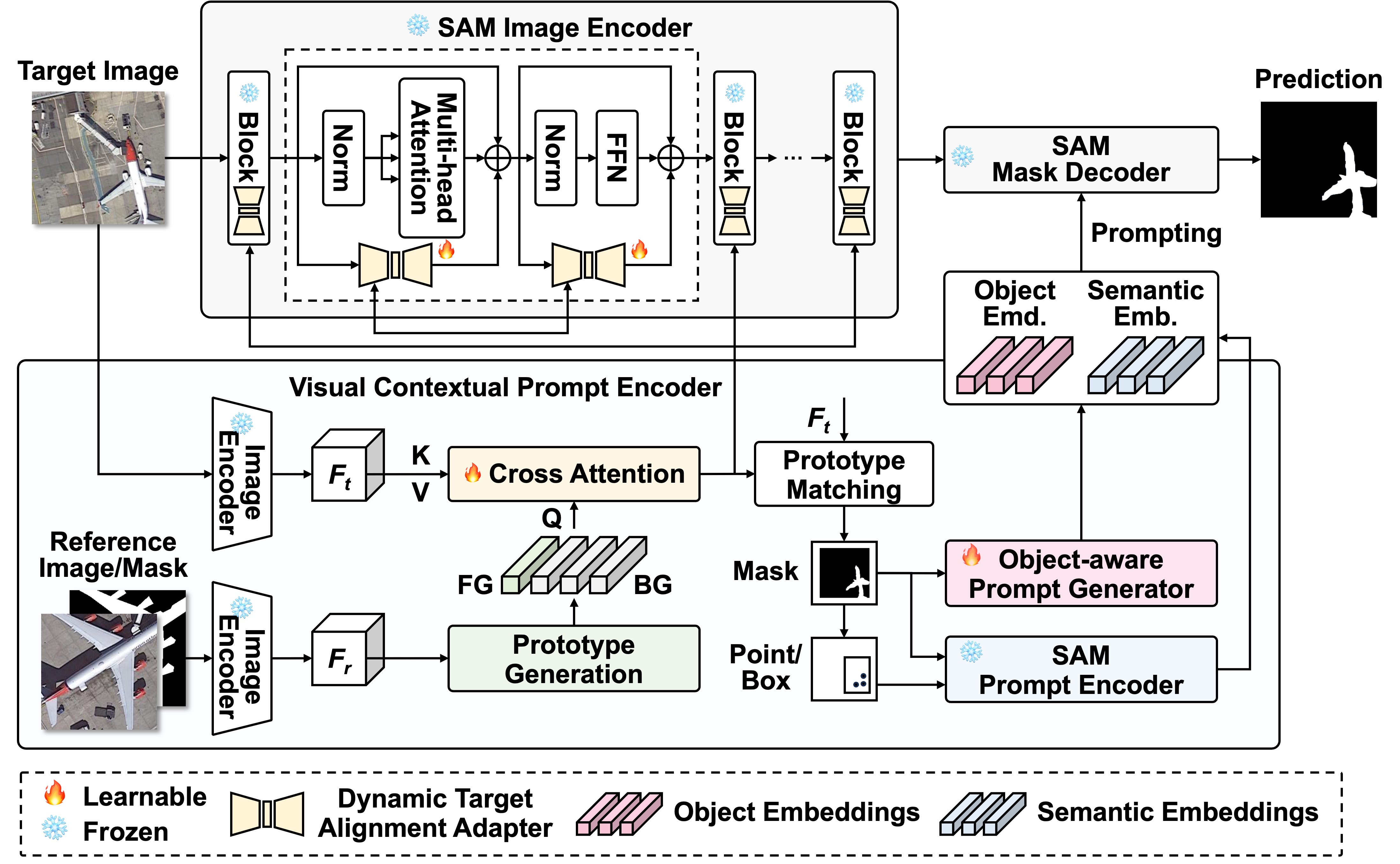}
\caption{\textbf{Overall framework of the proposed method ViRefSAM.} Without altering the original architecture of SAM, ViRefSAM introduces the reference image annotated with a specific object category and employs a Visual Contextual Prompt (VCP) Encoder to extract object-aware embeddings, which are injected into SAM’s mask decoder to guide segmentation. In parallel, a Dynamic Target Alignment (DTA) Adapter is integrated into the SAM image encoder to embed class-specific semantics from the reference image into the target image features, enabling SAM to dynamically focus on task-relevant objects. With this design, SAM can automatically segment all instances of the same category across different RS images using only a single reference image, eliminating the need for manual prompt customization.}
\label{fig:4}
\end{figure*}

\section{Problem Definition}\label{Problem Definition}
Existing SAM requires manual prompt customization for each image when segmenting specific objects, which becomes time-consuming and labor-intensive in complex remote sensing scenarios. To address this, inspired by few-shot segmentation (FSS), which utilizes a few labeled samples to guide the model in segmenting novel classes, we enhance SAM by incorporating reference images of specific categories. This enables SAM to automatically segment the same class of objects across multiple target images without needing manual prompts for each image. Therefore, we adopt the task definition from FSS. Following existing FSS methods~\cite{bi2023not,lang2024few}, we employ the meta-learning paradigm to train the model. Concretely, given a training dataset \emph{D}$_{train}$ containing seen classes \emph{C}$_{seen}$ and a testing dataset \emph{D}$_{test}$ with unseen classes \emph{C}$_{unseen}$ (where $C_{seen} \cap C_{unseen} = \emptyset$), the model learns transferable meta-knowledge from a series of tasks sampled from \emph{D}$_{train}$ and generalizes directly to \emph{D}$_{test}$. Both datasets \emph{D}$_{train}$ and \emph{D}$_{test}$ consist of multiple episodes, each containing a reference set $ R =\left \{ \left ( X^i_r, M^i_r\right )  \right \}_{i=1}^{K} $ and a target set $ T =\left \{ \left ( X_t, M_t\right )  \right \} $. Here, $X{}^{i}$ and $M_{}^{i}$ represent the original image and the binary mask containing the target class, respectively, and $K$ is the number of reference images provided in $R$. During each training episode, the model is optimized by predicting the target image $X_t$ based on the reference set $R$. After training, the model is evaluated on the testing episodes in \emph{D}$_{test}$ without further optimization, with the target mask $M_t$ in \emph{T} not being provided during the testing phase.

\noindent \textbf{Reference image annotation diversity:} To ensure annotation diversity, the reference image annotation $M_r$ supports multiple input forms, including points, scribbles, bounding boxes, and masks. Point annotations provide basic localization by placing a single point on the target object. Scribbles allow users to draw freehand curves within the object, offering flexibility in defining boundaries. Bounding boxes enclose the object with a rectangle, enabling quick and rough localization, especially in large or cluttered scenes. Masks provide pixel-level annotations for precise object delineation, ensuring high segmentation accuracy. These diverse annotation types enable our SAM to utilize varying degrees of reference images, balancing efficiency and accuracy according to task requirements.

\section{Proposed Method}\label{Proposed Method}
\subsection{Method Overview}

To address the high cost of manual prompt customization in SAM for RS object segmentation, we propose ViRefSAM, a novel framework that augments SAM with visual reference images containing class-specific objects, without modifying its core architecture. Under the guidance of these reference images, ViRefSAM enables automatic segmentation of target objects in RS images, eliminating the need to manually craft prompts for each image. Specifically, we introduce a Visual Contextual Prompt (VCP) Encoder that transforms the reference image into object-aware embeddings, which are then injected into SAM’s mask decoder to guide the segmentation of the desired object (see Fig.\ref{fig:4}). To further enhance SAM's task adaptability in RS scenarios, we incorporate a Dynamic Target Alignment (DTA) Adapter into the SAM image encoder. This adapter injects class-specific semantic clues from the reference image into the encoded features of the target image, enabling SAM to dynamically focus on task-relevant objects. With this design, ViRefSAM requires only a single reference image from a given category to automatically segment all instances of that category across diverse RS images, significantly improving segmentation efficiency and reducing manual effort.

The subsequent sections detail the core components of ViRefSAM: Section~\ref{Visual Contextual Prompt Encoder} presents the Visual Contextual Prompt Encoder, Section~\ref{Dynamic Target Alignment Adapter} introduces the Dynamic Target Alignment Adapter, and Section~\ref{ViRefSAM Prediction} outlines the segmentation prediction process and loss function design for model training.

\subsection{Visual Contextual Prompt Encoder} \label{Visual Contextual Prompt Encoder}

As depicted in Fig.\ref{fig:4}, we propose incorporating visual reference images containing specific object categories, which, under their guidance, drive SAM to automatically segment all instances of the same category across different RS images. To achieve this, we design a Visual Contextual Prompt (VCP) Encoder that encodes the reference images into class-specific embeddings, which are then input into SAM’s mask decoder to generate segmentation results. Specifically, the process consists of two steps: visual contextual interaction and object-aware prompt generation. The former involves interacting the reference image with the target image to generate a pseudo-mask for the target image, while the latter utilizes the generated pseudo-mask to produce the object-aware prompts, which are then fed into the SAM mask decoder.

\noindent \textbf{Visual Contextual Interaction.} Given the target image $X_t \in \mathbb{R}{^{3\times h \times w}}$ and the reference image $X_r \in \mathbb{R}{^{3\times h \times w}}$, both are fed into the frozen image encoder to generate the target features $F_t \in \mathbb{R}{^{C\times H \times W}}$ and reference features $F_r \in \mathbb{R}{^{C\times H \times W}}$, respectively. Next, to derive the class-specific semantics from the reference features, we perform the prototype generation process. Specifically, we separate the foreground and background regions of the reference image utilizing its mask. Given the complex and fragmented nature of backgrounds in RS scenarios, we adopt a Voronoi-based method~\cite{aurenhammer1991voronoi} to divide the background into multiple local regions, resulting in one foreground region $R_f \in \left \{ 0,1 \right \}^{H\times W}$ and several background local regions $\left \{ R_b^i \right \}_{i=1}^{N_b} $. Consequently, through masked average pooling, we can obtain both the foreground prototype $P_f \in \mathbb{R}{^{C\times 1}}$ and background prototypes $\left \{ P_b^i \right \}_{i=1}^{N_b} \in \mathbb{R}{^{C\times {N_b}}}$ of the reference image:
\begin{equation}\label{equation:1}
\setlength{\abovecaptionskip}{1pt}
\setlength{\belowcaptionskip}{1pt}
\begin{split}
P_f &= \frac{1}{\left | R_f \right | } \sum_{n=1}^{H\times W} F_{r;n} R_{f;n} \\
P_b^i &= \frac{1}{\left | R_{b}^i \right | } \sum_{n=1}^{H\times W} F_{r;n} R_{b;n}^i, \quad i=1,2,\dots,{N_b}
\end{split}
\end{equation}
Where $F_r;n$ denotes the $n^{th}$ pixel feature of the reference features. $R_b^i$ and $P_b^i$ represent the $i^{th}$ background local region and generated background prototype of the reference image, respectively.

To further capture class-specific clues from the target image itself, we utilize the reference prototypes to guide the mining process, as they already contain certain class-specific semantics. Specifically, we employ a cross-attention mechanism to aggregate semantics from the target image, where the reference prototypes $P \in \mathbb{R}{^{C\times \left({N_b}+1\right)}}$ serve as the query, and the target features $F_t \in \mathbb{R}{^{C\times\left(HW\right)}}$ act as the key and value:

\begin{align} \label{equation:2}
\setlength{\abovecaptionskip}{1pt}
\setlength{\belowcaptionskip}{1pt}
\widetilde{P}= \mathrm{Softmax}\left ( \frac{\left ( PW_q \right )\left ( F_tW_k \right )^T }{\sqrt{d} } \right ) \left ( F_tW_v \right )   
\end{align}
Where $\widetilde{P}=\left [\widetilde{P}_f,\widetilde{P}_b^1,...,\widetilde{P}_b^{N_b}\right ] $ denote the aligned prototypes, $\widetilde{P}^i_b$ denotes the $i^{th}$ background prototype, $F_t$ denotes the target image features and $W_*$ denote the projection weights. 

\begin{figure}[t]
\setlength{\abovecaptionskip}{2pt}
\centering
\includegraphics[width=1.0\linewidth]{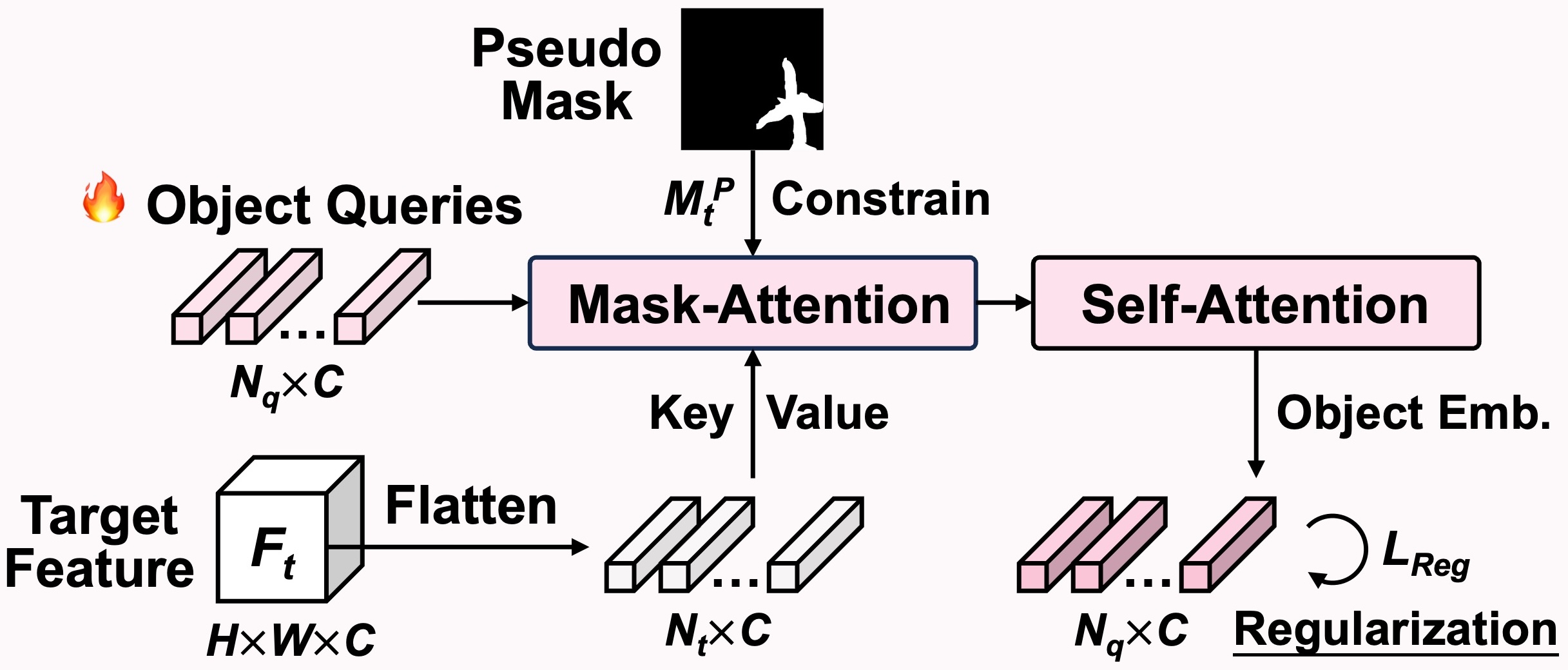}
\caption{\textbf{Detailed pipeline of the Object-aware Prompt Generator.} A set of learnable queries is initialized and enhanced by the reference prototype, then refined through mask attention under pseudo-mask constraints to extract object-specific semantics from the target image. A self-attention layer further aligns these embeddings with SAM’s feature space. To improve quality and avoid semantic redundancy, a regularization loss is applied to encourage diversity among the generated embeddings.}
\label{fig:5}
\end{figure}

After incorporating the class-specific semantics from the target image, we perform prototype matching by computing the similarity between the target image features $F_t$ and the reference prototypes $\widetilde{P}$ (which include a foreground prototpye $\widetilde{P}_f$ and ${N_b}$ background prototpyes $\left \{ \widetilde{P}_b^i \right \}_{i=1}^{N_b}$)to obtain its pseudo-mask $M_t^P \in \mathbb{R}{^{1\times H\times W}}$, which can be formulated as:

\begin{align}  \label{equation:3}
\setlength{\abovecaptionskip}{1pt}
\setlength{\belowcaptionskip}{1pt}
M_t^P=\left [\underset{i}{\operatorname{Max}}\,  \left (\bigcup_{i=1}^{N_b}\left \langle F_t, \widetilde{P}_b^{i} \right \rangle \right ); \left \langle F_t,\widetilde{P}_f \right \rangle 
\right ] 
\end{align}
Where $\left \langle \cdot,\cdot  \right \rangle $ denotes the similarity computation and $\bigcup $ denotes the concatenation of multiple segmentation results. Notably, the max operation $\mathrm{Max} \left(\cdot \right)$ is utilized to identify the background local prototype most similar to the target image pixels, effectively distinguishing between foreground and background pixels in the target image. Next, we feed the pseudo-mask of the target image into the object-aware prompt generator to generate a set of class-specific prompt embeddings.

\noindent \textbf{Object-aware Prompt Generation}, aims to produce a set of prompt embeddings for the SAM mask decoder, guided by the visual reference image. As illustrated in Fig.\ref{fig:5}, we introduce a set of learnable queries $Q \in \mathbb{R}^{C \times N_q}$, which are initialized utilizing Xavier initialization (with zero mean). These queries are further enhanced by adding the foreground prototype $\widetilde{P}_f$, thereby injecting class-specific semantic clues into the query representation. Subsequently, the queries perform mask attention under the constraint of the pseudo mask $M_t^P$ derived from the visual contextual interaction, enabling them to absorb additional class-specific information from the target image $X_t$. Finally, a self-attention layer is applied to refine the queries and align them with the SAM representation space, yielding a set of object-aware prompt embeddings suitable for guiding segmentation. The entire process can be formulated as:

\begin{align}  \label{equation:4}
\setlength{\abovecaptionskip}{1pt}
\setlength{\belowcaptionskip}{1pt}
P_o = \mathrm{SelfAttn}\left ( \mathrm{MaskAttn\left ( Q, F_t, M_t^P \right ) }  \right ) 
\end{align}
Where $P_o \in \mathbb{R}^{C \times N_q}$ denotes the object-aware prompt embeddings. 

To improve the quality of object-aware prompt embeddings and avoid semantic redundancy, we introduce a regularization loss $L_{reg}$ that encourages these embeddings to be mutually decorrelated while maintaining consistency in representing the same target object. Specifically, given the prompt embedding set $\{P_o^i\}_{i=1}^{N_q} \in \mathbb{R}^C$, the loss is defined as:
\begin{align}  \label{equation:5}
\setlength{\abovecaptionskip}{1pt}
\setlength{\belowcaptionskip}{1pt}
{ \mathcal{L}_{reg}} =\frac{\sum_{i=1}^{N_q}\sum_{j\neq i\ }^{N_q} { \left \langle P_o^i, P_o^j \right \rangle   } }{N_q*\left(N_q-1\right)} 
\end{align}
Where $\left\langle \cdot, \cdot \right\rangle$ denotes the inner product between prompt embeddings. This soft regularization encourages the embeddings to be mutually distinctive while remaining semantically aligned with the same object category. By promoting diversity without enforcing strict orthogonality, it mitigates redundancy and enhances the expressiveness of the prompt representation, ultimately leading to more accurate and robust segmentation.

Additionally, to further enhance the diversity and expressive power of the prompts, besides the proposed object-aware embeddings, we also leverage SAM’s own prompt encoder to extract semantic embeddings from the pseudo-mask and its converted points and bounding boxes. These semantic embeddings $P_s$ are jointly fed into SAM’s mask decoder along with the object-aware embeddings to facilitate accurate target segmentation.

\subsection{Dynamic Target Alignment Adapter} \label{Dynamic Target Alignment Adapter}

Through the design in Sec.\ref{Visual Contextual Prompt Encoder}, ViRefSAM significantly improves SAM’s efficiency in RS object segmentation. However, since SAM’s pretraining largely relies on natural scene data, it lacks domain-specific knowledge for RS tasks, resulting in lower accuracy for segmenting RS objects. To address this, we propose a Dynamic Target Alignment (DTA) Adapter. This adapter introduces an additional structure in the SAM image encoder and is fine-tuned separately to inject RS domain knowledge while maintaining SAM’s image representation capability. 

\begin{figure}[t]
\setlength{\abovecaptionskip}{2pt}
\centering
\includegraphics[width=1.0\linewidth]{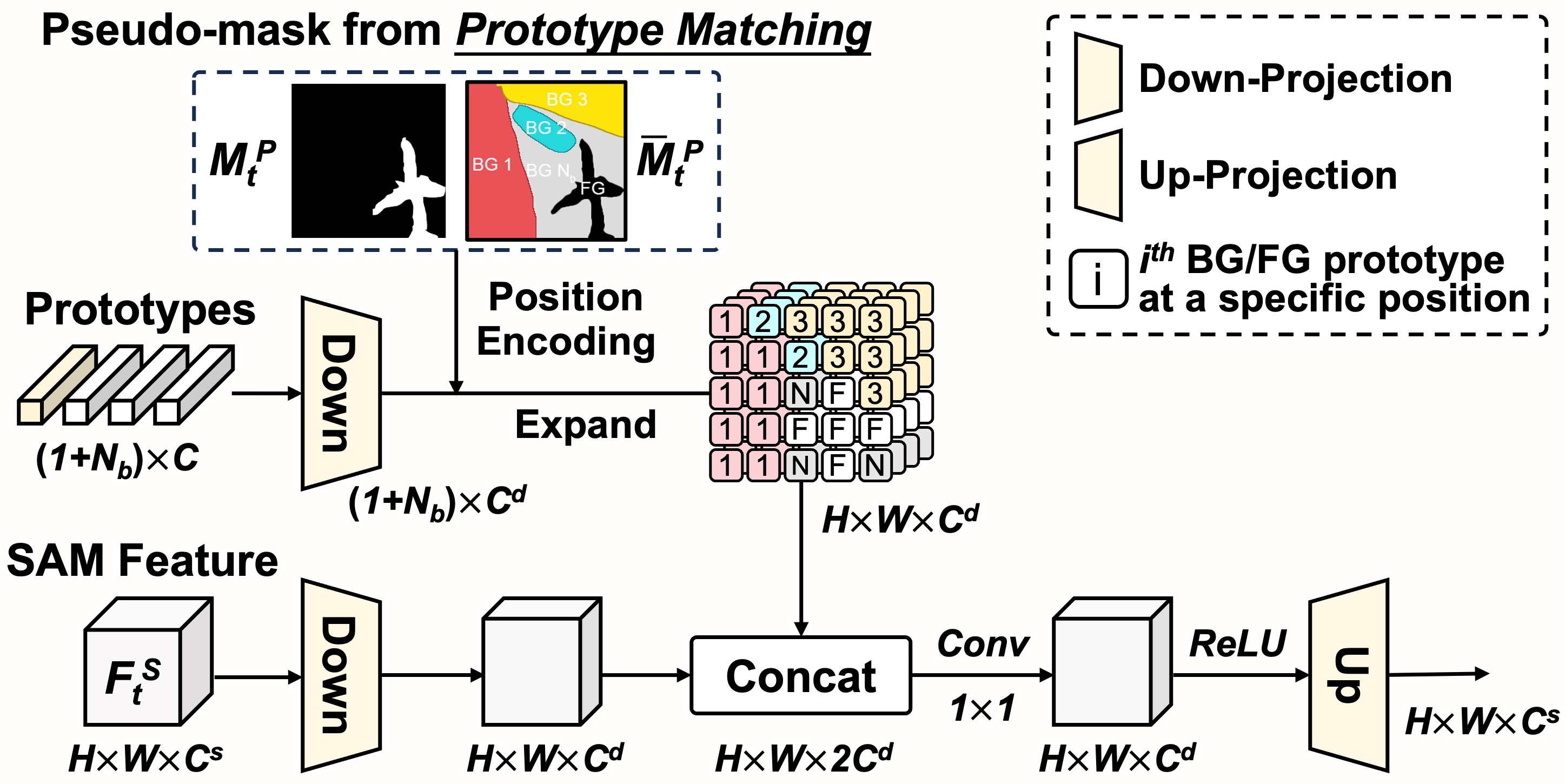}
\caption{\textbf{Detailed pipeline of the Dynamic Target Alignment (DTA) Adapter.} The DTA adapter injects class-specific semantics from the reference image into the encoded features of the target image under positional constraints. This is achieved by embedding the reference prototypes into the target features, enabling SAM to dynamically attend to task-relevant objects during segmentation.}
\label{fig:6}
\end{figure}

Specifically, as illustrated in Fig.\ref{fig:4}, we insert a DTA adapter in parallel within each block of the SAM image encoder. During training, the parameters of SAM’s original image encoder are frozen, and only the parameters of the adapter are fine-tuned. Given the input target image token $x$, the output of each block can be expressed as:
\begin{align}  \label{equation:6}
\setlength{\abovecaptionskip}{1pt}
\setlength{\belowcaptionskip}{1pt}
\mathrm{Block}\left ( x \right )=\mathrm{MultiAttn}\left ( \mathrm{Norm} \left ( x \right )  \right ) +x+\mathrm{DTA}\left ( x \right )  
\end{align}
Where $\mathrm{MultiAttn}\left (\cdot \right )$ and $\mathrm{Norm}\left (\cdot \right )$ are the original structures in the encoder, and $\mathrm{DTA}\left (\cdot \right )$ is the proposed adapter structure. Next, we provide a detailed introduction to the DTA adapter. 

As depicted in Fig.\ref{fig:6}, for the features $F_t^S \in \mathbb{R}^{C^s\times H \times W}$ of the target image entering each block, they are first mapped to a $C^d$-dimensional space via a down-projection layer. Unlike original adapter structures, we extract class-specific clues from the reference image prototypes $\widetilde{P}  \in \mathbb{R}^{C\times \left(1+N_b\right) }$ and inject them into SAM’s features to dynamically focus on the relevant objects for the current task. Specifically, the reference prototypes are also mapped to the $C^d$-dimensional space utilizing the down-projection layer. The prototypes are then expanded to match the size of the image features for feature concatenation.

To achieve this, we design a dedicated position encoding that ensures the alignment of reference prototypes $\widetilde{P}$ with specific locations in the target image. Since the pseudo-background mask $\bar{M}_t^P$ of the target image, as defined in Eq. (\ref{equation:3}), is derived from the maximum similarity score between multiple reference background prototypes, we can determine the background or foreground prototype ID for each pixel. This guarantees that during the expansion process, different reference prototypes are assigned to the correct pixel positions. This encoding method allows the model to effectively distinguish between foreground and background regions, enhancing segmentation accuracy and ensuring that the appropriate prototype influences the corresponding regions in the image.

After feature concatenation, we utilize a 1$\times$1 convolution to perform feature fusion and apply an up-projection layer to restore the feature dimensions to the $C^s$-dimensional space, generating the final output of the DTA adapter. The entire process can be formalized as:
\begin{align}  \label{equation:7}
\setlength{\abovecaptionskip}{1pt}
\setlength{\belowcaptionskip}{1pt}
 \mathrm{DTA}\left ( F_t^S \right ) =\left ( \mathrm{Conv}\left [ F_t^SW_{down}; \mathrm{Pos}\left (\widetilde{P}W_{down}  \right )   \right ]  \right )W_{up} 
\end{align}
Where $W_{down}\left ( \cdot \right )$ and $W_{up}\left ( \cdot \right )$ represent the down- and up-projection layers, respectively. $\mathrm{Conv}\left ( \cdot \right )$ refers to the 1$\times$1 convolution operation utilized for feature fusion, and $\mathrm{Pos}\left ( \cdot \right ) $ denotes the position encoding process, which aligns the reference prototypes with specific locations in the target image.

\begin{align}  \label{equation:8}
\setlength{\abovecaptionskip}{1pt}
\setlength{\belowcaptionskip}{1pt}
 \mathrm{Pos}\left (\widetilde{P}W_{down}  \right )\left ( i,j \right )=\begin{cases}
  \widetilde{P}^mW_{down}, & \text{ if }  \bar{M}^P_t(i,j)=m  \\
  0,& \text{else}
\end{cases}
\end{align}
Here, we present the specific formula for the position encoding process that we have developed, where $m$ denotes the $m^{th}$ reference prototype from $\widetilde{P}  \in \mathbb{R}^{C\times \left(1+N_b\right) }$.

\subsection{ViRefSAM Prediction} \label{ViRefSAM Prediction}

\noindent \textbf{1-shot Setting.} Overall, given a target image $X_t$ and a reference image $X_r$ with its associated annotation $M_r$, the segmentation result $\mathcal{M}_t$ of ViRefSAM can be formulated as:
\begin{equation}\label{equation:9}
\setlength{\abovecaptionskip}{1pt}
\setlength{\belowcaptionskip}{1pt}
\begin{split}
\mathcal{M}_t&=\mathrm{ViRefSAM}\left(X_t,X_r,M_r \right)
\\
&= \mathrm{SAM Dec} \left ( \mathrm{SAM Enc}\left ( X_t \right ), \mathrm{VCPEnc}\left ( X_t,X_r,M_r \right )   \right ) 
\end{split}
\end{equation}
Where $\mathrm{SAMEnc}(\cdot)$ represents the SAM image encoder equipped with the proposed Dynamic Target Alignment Adapter, $\mathrm{VCPEnc}(\cdot)$ denotes the Visual Contextual Prompt Encoder that generates object-aware embeddings based on both the target and reference inputs, and $\mathrm{SAMDec}(\cdot)$ refers to the original SAM mask decoder. This formulation enables the model to generate accurate segmentation masks $\mathcal{M}_t$ under visual contextual guidance without manual prompt design.

\noindent \textbf{K-shot Setting.} When extending to the K-shot setting (i.e., K$>$1), multiple reference images are provided to guide the segmentation. Our ViRefSAM can effectively handle this case by feeding each reference image and its corresponding mask into the VCP Encoder to generate a pair of object-aware prompt embeddings and semantic embeddings. This results in K pairs of embeddings, which are collectively input into SAM’s mask decoder to produce the segmentation output. This flexible design benefits from the inherent self-attention mechanism of the SAM architecture.

\noindent \textbf{Loss Function.} To supervise the training of ViRefSAM, we adopt a combination of Binary Cross-Entropy (BCE) loss and Dice loss. The BCE loss ensures pixel-wise accuracy, while the Dice loss complements it by capturing region-level consistency. The overall loss function of ViRefSAM is defined as:
\begin{align}  \label{equation:10}
\setlength{\abovecaptionskip}{1pt}
\setlength{\belowcaptionskip}{1pt}
\mathcal{L}= \mathrm{BCE}\left ( \mathcal{M}_t,M_t  \right )  + \mathrm{Dice}\left ( \mathcal{M}_t,M_t  \right )+\gamma \mathcal{L}_{reg} 
\end{align}
Where $\gamma$ denotes the contribution of the $\mathcal{L}_{reg}$.

\begin{figure*}[t]
\setlength{\abovecaptionskip}{1pt} \setlength{\belowcaptionskip}{1pt}
\centering
\includegraphics[width=1.0\linewidth]{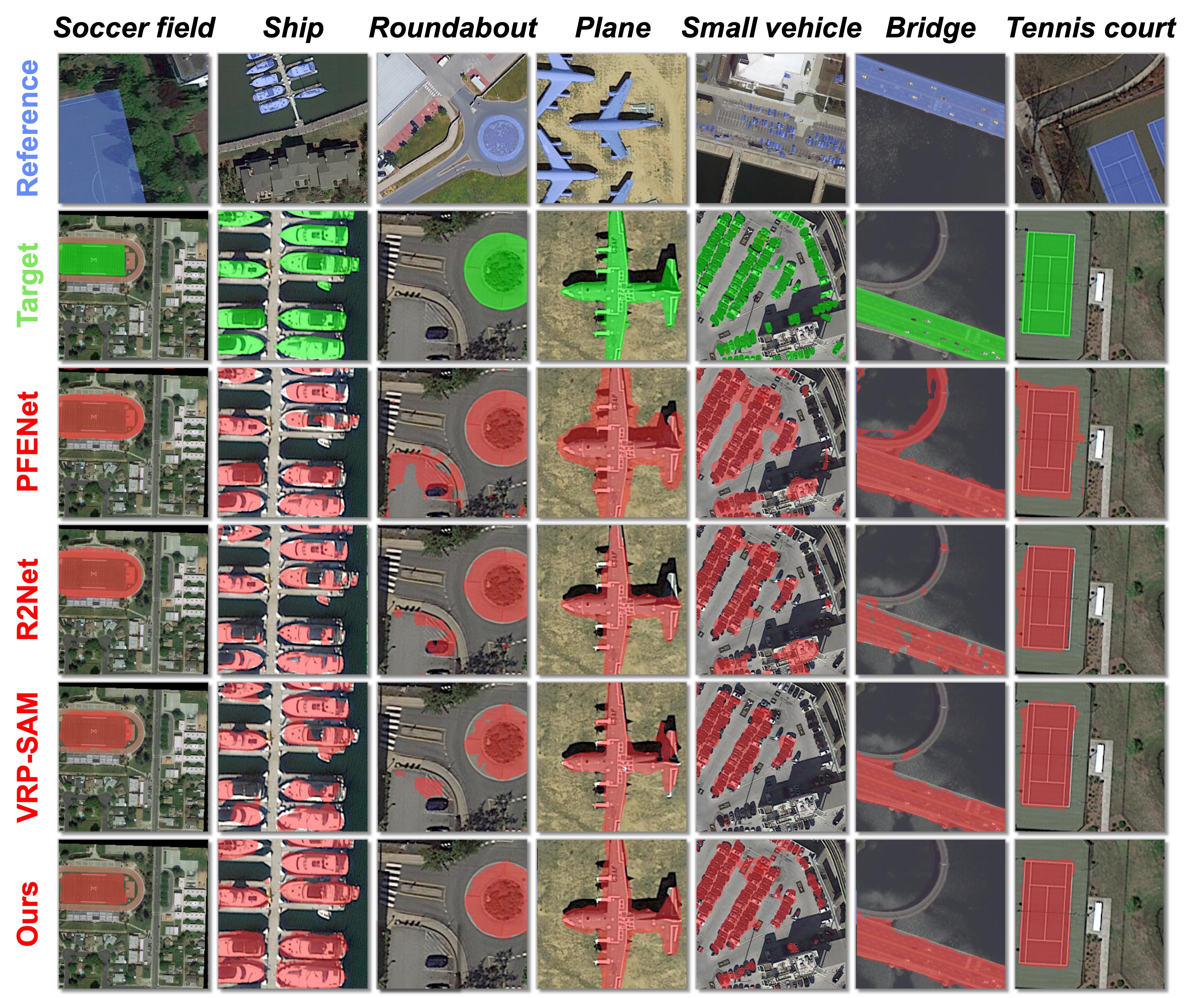}
\caption{\textbf{Qualitative visualization of our ViRefSAM and several competing methods under the 1-shot setting on iSAID-5$^i$ benchmark.} From top to bottom, each row shows: (1) reference images with ground truth masks; (2) target images with ground truth masks; (3) segmentation results of R2Net; (4) segmentation results of VRP-SAM; and (5) segmentation results of our ViRefSAM.}
\label{fig:7}
\end{figure*}

\begin{figure}[t]
\setlength{\abovecaptionskip}{1pt} \setlength{\belowcaptionskip}{1pt}
\centering
\includegraphics[width=1.0\linewidth]{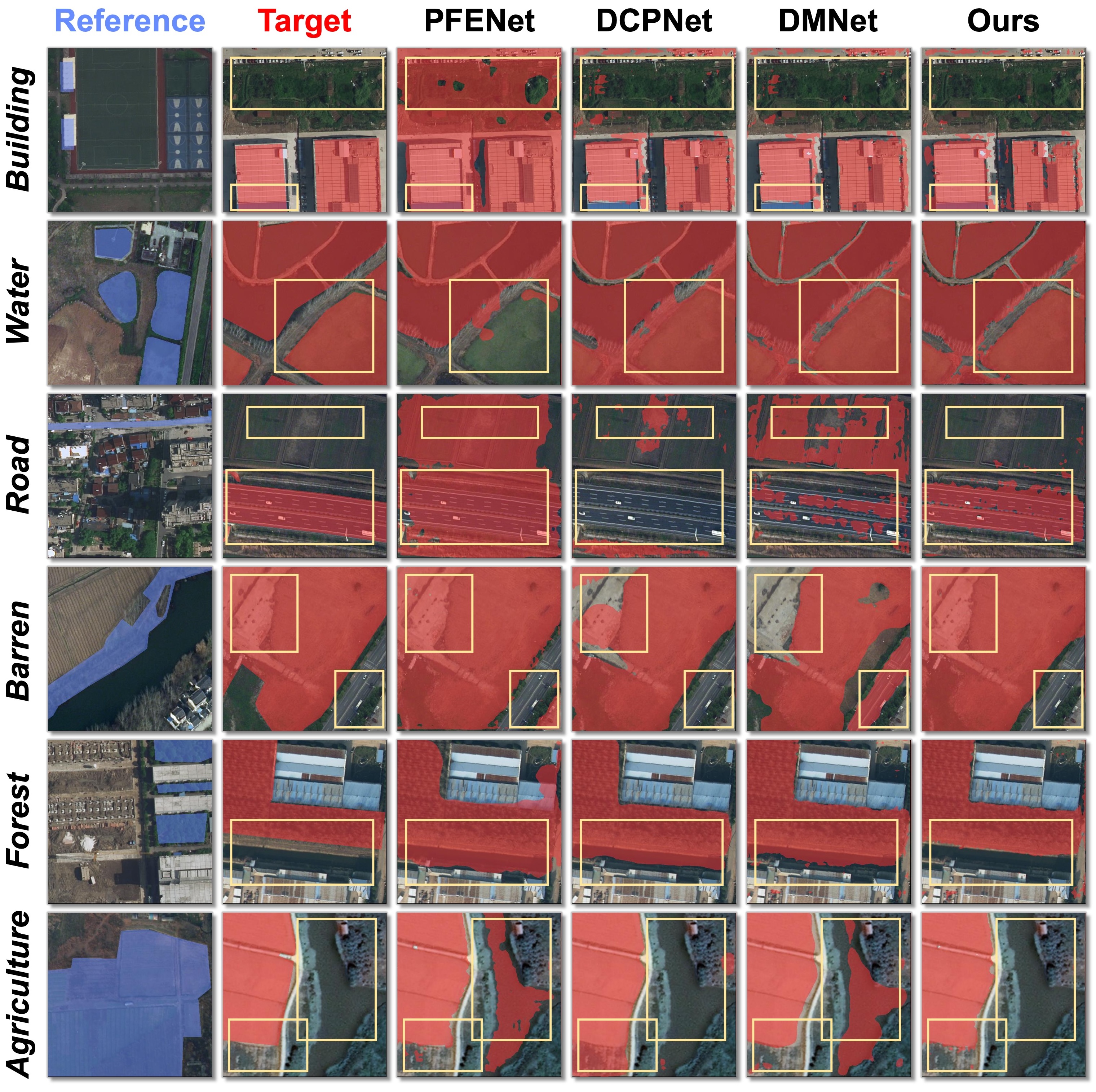}
\caption{\textbf{Qualitative visualization of our ViRefSAM and several competing methods under the 1-shot setting on LoveDA-2$^i$ benchmark.} From left to right, each column shows: (1) reference images with ground truth masks; (2) target images with ground truth masks; (3) segmentation results of PFENet; (4) segmentation results of DCPNet; (5) segmentation results of DMNet; and (6) segmentation results of our ViRefSAM.}
\label{fig:8}
\end{figure}

\begin{figure}[t]
\setlength{\abovecaptionskip}{1pt} \setlength{\belowcaptionskip}{1pt}
\centering
\includegraphics[width=1.0\linewidth]{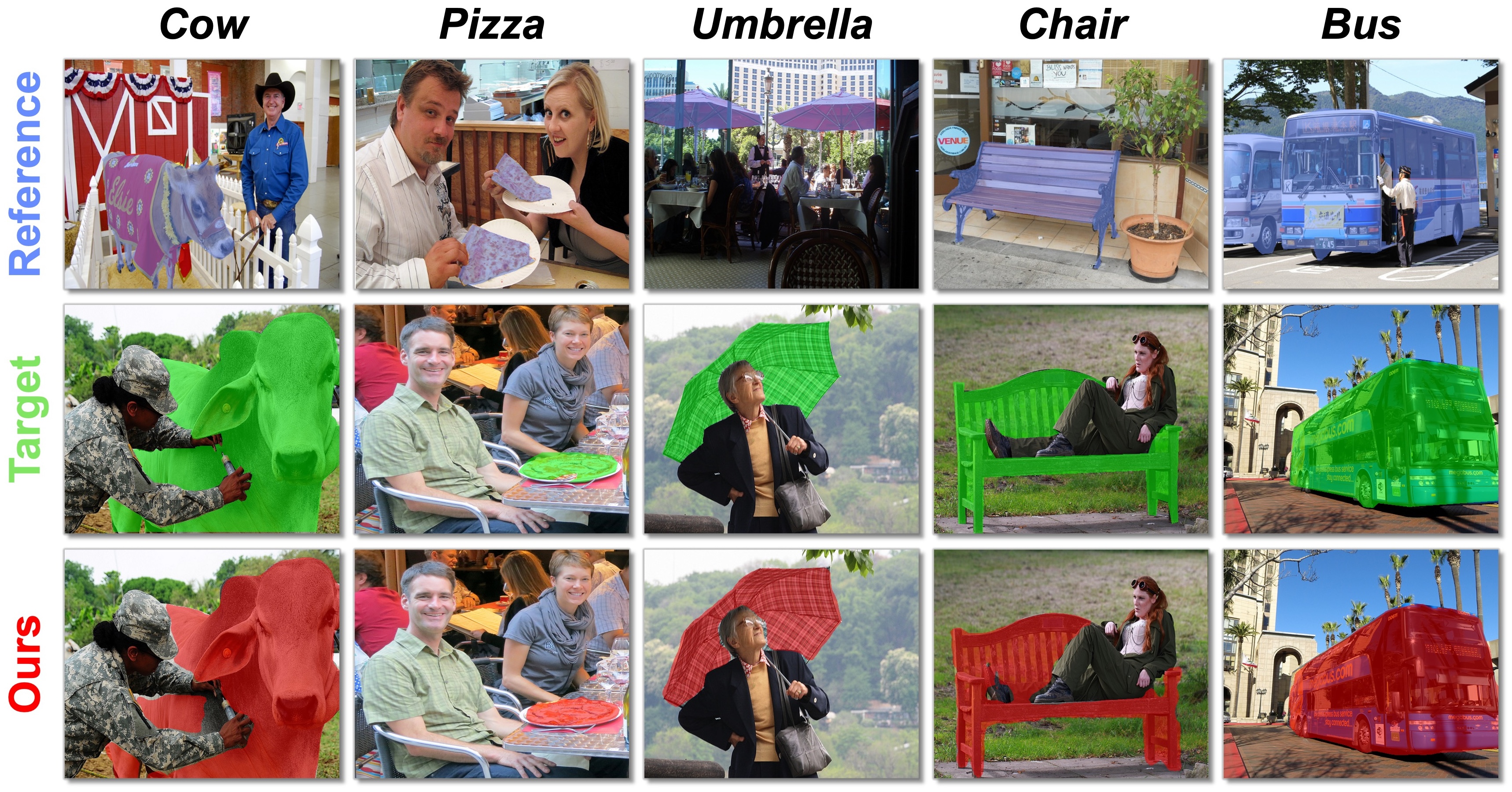}
\caption{\textbf{Qualitative visualization of our ViRefSAM under the 1-shot setting on COCO-20$^i$ benchmark.} From top to bottom, each row shows: (1) reference images with ground truth masks; (2) target images with ground truth masks; (3) segmentation results of our ViRefSAM.}
\label{fig:9}
\end{figure}

\begin{table*}[t]
\setlength{\abovecaptionskip}{5pt}
\setlength{\belowcaptionskip}{10pt}
\caption{ Performance comparison on iSAID-5$^i$ under the 1-shot and 5-shot settings, utilizing mIoU and FB-IoU (\%) as evaluation metrics. ``Mean” denotes the average mIoU across the three folds, and ``FB-IoU” represents the average foreground-background IoU over the same folds.} \label{tab:1}
\renewcommand\arraystretch{1.25}
\centering
\resizebox{1.0\linewidth}{!}{
\begin{tabular}{cr|ccccc|ccccc} 
\toprule
\multirow{2}{*}{Backbone}    & \multirow{2}{*}{Method}                                      & \multicolumn{5}{c|}{1-shot}                                                                                                                                                                                                                                            & \multicolumn{5}{c}{5-shot}                                                                                                                                                                                                                                              \\ 
\cline{3-12}
&                                                              & Fold-0                                             & Fold-1                                             & Fold-2                                             & Mean                                               & FB-IoU                                             & Fold-0                                             & Fold-1                                             & Fold-2                                             & Mean                                               & FB-IoU                                              \\ 
\hline
\multirow{12}{*}{ResNet-50}  & ASGNet (CVPR'21)\cite{li2021adaptive}                                             & 48.59                                              & 36.82                                              & 46.65                                              & 44.02                                              & 62.36                                              & 53.01                                              & 37.44                                              & 52.18                                              & 47.54                                              & 64.63                                               \\
& CyCTR (NeurIPS'21)\cite{zhang2021few}                                           & 51.15                                              & 38.40                                              & 53.79                                              & 47.78                                              & 62.86                                              & 51.91                                              & 39.01                                              & 54.83                                              & 48.58                                              & 63.81                                               \\
& PFENet (TPAMI'22)\cite{tian2022prior}                                            & 51.34                                              & 38.79                                              & 52.26                                              & 47.46                                              & 63.34                                              & 54.71                                              & 41.51                                              & 54.45                                              & 50.22                                              & 64.99                                               \\
& NERTNet (CVPR'22)\cite{liu2022learning}                                            & 49.52                                              & 38.66                                              & 51.87                                              & 46.68                                              & 62.84                                              & 50.60                                              & 40.99                                              & 55.07                                              & 48.89                                              & 63.74                                               \\
& DCPNet (IJCV'23)\cite{lang2024few}                                             & 48.43                                              & 37.59                                              & 52.09                                              & 46.04                                              & 62.76                                              & 50.34                                              & 40.75                                              & 51.33                                              & 47.47                                              & 64.10                                               \\
& SCCAN (ICCV'23)\cite{xu2023self}                                              & 52.24                                              & 38.60                                              & 52.84                                              & 47.89                                              & 63.26                                              & 52.75                                              & 38.95                                              & 54.26                                              & 48.65                                              & 62.23                                               \\
& BAM (TPAMI'23)\cite{lang2023base}                                               & 58.17                                              & 42.12                                              & 50.95                                              & 50.41                                              & 67.03                                              & 62.51                                              & 43.16                                              & 58.24                                              & 54.64                                              & 67.92                                               \\
& R2Net (TGRS'23)\cite{lang2023global}                                              & 56.81                                              & 39.85                                              & 49.02                                              & 48.56                                              & 62.65                                              & 60.47                                              & 41.43                                              & 50.24                                              & 50.71                                              & 65.70                                               \\
& DMNet (TGRS'23)\cite{bi2023not}                                              & 54.45                                              & 40.68                                              & 53.60                                              & 49.58                                              & 64.46                                              & 57.67                                              & 41.06                                              & 55.28                                              & 51.34                                              & 65.81                                               \\
& MGANet (TGRS'24)\cite{MGANet}                                             & 55.81                                              & 43.52                                              & 56.23                                              & 51.85                                              & 64.49                                              & 57.81                                              & 45.32                                              & 57.87                                              & 53.67                                              & 65.99                                               \\
& VRP-SAM (CVPR'24)\cite{sun2024vrp}                                            & 52.37                                              & 41.62                                              & 53.60                                              & 49.20                                              & 64.55                                              & 55.08                                              & 43.21                                              & 56.95                                              & 51.75                                              & 65.57                                               \\
& {\cellcolor[rgb]{0.941,0.941,0.941}}\textbf{ViRefSAM (Ours)} & {\cellcolor[rgb]{0.941,0.941,0.941}}\textbf{57.92} & {\cellcolor[rgb]{0.941,0.941,0.941}}\textbf{44.74} & {\cellcolor[rgb]{0.941,0.941,0.941}}\textbf{56.51} & {\cellcolor[rgb]{0.941,0.941,0.941}}\textbf{53.06} & {\cellcolor[rgb]{0.941,0.941,0.941}}\textbf{67.47} & {\cellcolor[rgb]{0.941,0.941,0.941}}\textbf{61.23} & {\cellcolor[rgb]{0.941,0.941,0.941}}\textbf{46.48} & {\cellcolor[rgb]{0.941,0.941,0.941}}\textbf{59.52} & {\cellcolor[rgb]{0.941,0.941,0.941}}\textbf{55.74} & {\cellcolor[rgb]{0.941,0.941,0.941}}\textbf{68.34}  \\ 
\hline
\multirow{12}{*}{ResNet-101} & ASGNet (CVPR'21)\cite{li2021adaptive}                                             & 47.55                                              & 38.47                                              & 49.28                                              & 45.10                                              & 62.03                                              & 53.54                                              & 38.24                                              & 53.20                                              & 48.33                                              & 65.35                                               \\
& CyCTR (NeurIPS'21)\cite{zhang2021few}                                           & 50.89                                              & 38.89                                              & 52.22                                              & 47.33                                              & 62.35                                              & 52.15                                              & 40.28                                              & 55.32                                              & 49.25                                              & 64.45                                               \\
& PFENet (TPAMI'22)\cite{tian2022prior}                                            & 50.69                                              & 38.37                                              & 52.85                                              & 47.30                                              & 62.46                                              & 54.40                                              & 41.55                                              & 50.55                                              & 48.83                                              & 64.57                                               \\
& NERTNet (CVPR'22)\cite{liu2022learning}                                            & 50.33                                              & 38.73                                              & 51.23                                              & 46.76                                              & 63.25                                              & 53.24                                              & 41.87                                              & 51.53                                              & 48.88                                              & 64.16                                               \\
& DCPNet (IJCV'23)\cite{lang2024few}                                             & 47.63                                              & 38.80                                              & 49.34                                              & 45.26                                              & 62.56                                              & 50.68                                              & 40.02                                              & 54.52                                              & 48.41                                              & 62.96                                               \\
& SCCAN (ICCV'23)\cite{xu2023self}                                              & 52.71                                              & 38.76                                              & 53.45                                              & 48.31                                              & 62.72                                              & 53.85                                              & 39.11                                              & 54.56                                              & 49.17                                              & 64.27                                               \\
& BAM (TPAMI'23)\cite{lang2023base}                                               & 59.10                                              & 41.65                                              & 50.54                                              & 50.43                                              & 66.05                                              & 61.57                                              & 42.32                                              & 55.34                                              & 53.08                                              & 66.42                                               \\
& R2Net (TGRS'23)\cite{lang2023global}                                              & 57.74                                              & 39.76                                              & 48.45                                              & 48.65                                              & 63.38                                              & 58.83                                              & 40.58                                              & 49.69                                              & 49.70                                              & 63.79                                               \\
& DMNet (TGRS'23)\cite{bi2023not}                                              & 54.01                                              & 40.04                                              & 53.57                                              & 49.21                                              & 64.03                                              & 55.70                                              & 41.69                                              & 56.47                                              & 51.29                                              & 65.88                                               \\
& MGANet (TGRS'24)\cite{MGANet}                                             & 55.03                                              & 42.90                                              & 54.69                                              & 50.87                                              & 64.35                                              & 56.43                                              & 44.85                                              & 57.67                                              & 52.98                                              & 66.42                                               \\
& VRP-SAM (CVPR'24)\cite{sun2024vrp}                                            & 51.68                                              & 41.88                                              & 53.52                                              & 49.03                                              & 65.41                                              & 55.36                                              & 44.09                                              & 56.31                                              & 51.92                                              & 65.13                                               \\
& {\cellcolor[rgb]{0.941,0.941,0.941}}\textbf{ViRefSAM (Ours)} & {\cellcolor[rgb]{0.941,0.941,0.941}}\textbf{57.10} & {\cellcolor[rgb]{0.941,0.941,0.941}}\textbf{44.43} & {\cellcolor[rgb]{0.941,0.941,0.941}}\textbf{56.84} & {\cellcolor[rgb]{0.941,0.941,0.941}}\textbf{52.79} & {\cellcolor[rgb]{0.941,0.941,0.941}}\textbf{67.22} & {\cellcolor[rgb]{0.941,0.941,0.941}}\textbf{60.38} & {\cellcolor[rgb]{0.941,0.941,0.941}}\textbf{46.71} & {\cellcolor[rgb]{0.941,0.941,0.941}}\textbf{59.22} & {\cellcolor[rgb]{0.941,0.941,0.941}}\textbf{55.44} & {\cellcolor[rgb]{0.941,0.941,0.941}}\textbf{68.87}  \\ 
\hline
\multirow{3}{*}{ViT-B/16}    & FPTrans (NeurIPS'22)\cite{zhang2022feature}                                         & 50.04                                              & 36.85                                              & 53.30                                              & 46.73                                              & 62.36                                              & 57.92                                              & 47.42                                              & 64.23                                              & 56.52                                              & 66.12                                               \\
& AgMTR (IJCV'24)\cite{bi2024agmtr}                                              & 55.45                                              & 40.67                                              & 53.81                                              & 49.98                                              & 63.59                                              & 59.05                                              & 47.86                                              & 64.08                                              & 57.00                                              & 67.60                                               \\
& {\cellcolor[rgb]{0.941,0.941,0.941}}\textbf{ViRefSAM (Ours)} & {\cellcolor[rgb]{0.941,0.941,0.941}}\textbf{56.34} & {\cellcolor[rgb]{0.941,0.941,0.941}}\textbf{42.50} & {\cellcolor[rgb]{0.941,0.941,0.941}}\textbf{55.29} & {\cellcolor[rgb]{0.941,0.941,0.941}}\textbf{51.38} & {\cellcolor[rgb]{0.941,0.941,0.941}}\textbf{65.25} & {\cellcolor[rgb]{0.941,0.941,0.941}}\textbf{59.74} & {\cellcolor[rgb]{0.941,0.941,0.941}}\textbf{48.23} & {\cellcolor[rgb]{0.941,0.941,0.941}}\textbf{63.94} & {\cellcolor[rgb]{0.941,0.941,0.941}}\textbf{57.30} & {\cellcolor[rgb]{0.941,0.941,0.941}}\textbf{68.49}  \\ 
\hline
\multirow{3}{*}{DeiT-B/16}   & FPTrans (NeurIPS'22)\cite{zhang2022feature}                                         & 51.59                                              & 37.79                                              & 52.85                                              & 47.41                                              & 62.52                                              & 58.52                                              & 47.24                                              & 65.37                                              & 57.04                                              & 67.63                                               \\
& AgMTR (IJCV'24)\cite{bi2024agmtr}                                              & 57.51                                              & 43.54                                              & 53.69                                              & 51.58                                              & 64.63                                              & 61.51                                              & 49.23                                              & 64.05                                              & 58.26                                              & 68.72                                               \\
& {\cellcolor[rgb]{0.941,0.941,0.941}}\textbf{ViRefSAM (Ours)} & {\cellcolor[rgb]{0.941,0.941,0.941}}\textbf{58.13} & {\cellcolor[rgb]{0.941,0.941,0.941}}\textbf{45.26} & {\cellcolor[rgb]{0.941,0.941,0.941}}\textbf{56.01} & {\cellcolor[rgb]{0.941,0.941,0.941}}\textbf{53.13} & {\cellcolor[rgb]{0.941,0.941,0.941}}\textbf{66.99} & {\cellcolor[rgb]{0.941,0.941,0.941}}\textbf{62.20}  & {\cellcolor[rgb]{0.941,0.941,0.941}}\textbf{49.57} & {\cellcolor[rgb]{0.941,0.941,0.941}}\textbf{63.28} & {\cellcolor[rgb]{0.941,0.941,0.941}}\textbf{58.35} & {\cellcolor[rgb]{0.941,0.941,0.941}}\textbf{69.25}  \\
\bottomrule
\end{tabular}}
\end{table*}

\begin{table*}[t]
\setlength{\abovecaptionskip}{5pt}
\setlength{\belowcaptionskip}{10pt}
\caption{Fine-grained class-wise performance comparison on iSAID-5$^i$ under the 1-shot setting using IoU as the evaluation metric. C1–C15 denote the unseen classes across the three evaluation folds, where C1–C5 are from Fold 0, C6–C10 from Fold 1, and C11–C15 from Fold 2. ``Mean” denotes the average mIoU across the three folds.} \label{tab:2}
\renewcommand\arraystretch{1.25}
\centering
\resizebox{1.0\linewidth}{!}{
\begin{threeparttable}
\begin{tabular}{r|ccccccccccccccc|c} 
\toprule
Method                                            & C1             & C2             & C3             & C4             & C5             & C6             & C7             & C8             & C9             & C10            & C11            & C12            & C13            & C14            & C15            & Mean            \\ 
\hline
\multicolumn{17}{c}{ResNet-50}                                                                                                                                                                                                                                                                                                     \\ 
\hline
ASGNet (CVPR'21)\cite{li2021adaptive}                                  & 42.83          & 50.95          & \underline{61.00}  & 39.23          & 48.96          & 39.75          & 49.70          & 33.04          & 31.96          & 29.67          & \underline{72.48}  & 48.73          & 45.23          & 24.00          & \underline{42.83}  & 44.02           \\
CyCTR (NeurIPS'21)\cite{zhang2021few}                                & 52.20          & 51.41          & 60.14          & 40.01          & \textbf{51.97} & \textbf{45.52} & \underline{53.04}  & 35.23          & 29.89          & 28.34          & \textbf{76.77} & 49.51          & \underline{63.12}  & 21.59          & \textbf{57.98} & 47.78           \\
PFENet (TPAMI'22)\cite{tian2022prior}                                 & 50.77          & 47.05          & 53.69          & 63.03          & 42.13          & 23.84          & 48.61          & 47.96          & \underline{37.95}  & 35.57          & 50.98          & 65.33          & 55.98          & 65.79          & 23.22          & 47.46           \\
NERTNet (CVPR'22)\cite{liu2022learning}                                 & 49.46          & 52.34          & 43.42          & 61.62          & 40.76          & 27.87          & 45.48          & 48.87          & 35.12          & 35.96          & 52.78          & 60.71          & 57.11          & 63.95          & 24.8           & 46.68           \\
DCPNet (IJCV'23)\cite{lang2024few}                                  & 47.44          & 47.54          & 49.19          & 63.84          & 34.13          & 23.32          & 47.09          & 49.49          & 30.34          & \textbf{37.72} & 44.79          & 70.2           & 58.94          & 64             & 22.55          & 46.04           \\
SCCAN (ICCV'23)\cite{xu2023self}                                   & \textbf{54.03} & 52.15          & \textbf{63.41} & 55.14          & 36.47          & 32.59          & 43.50          & 52.82          & 30.43          & 33.64          & 49.60          & \underline{70.92}  & \textbf{65.22} & 58.50          & 19.98          & 47.89           \\
BAM (TPAMI'23)\cite{lang2023base}                                    & 51.43          & \underline{64.15}  & 60.98          & 65.08          & 49.21          & 33.64          & 50.35          & \underline{56.81}  & 33.47          & \underline{36.33}  & 50.97          & 64.79          & 50.67          & \textbf{67.48} & 20.84          & \underline{50.41}   \\
R2Net (TGRS'23)\cite{lang2024few}                                   & 45.25          & 62.54          & 58.59          & \textbf{67.54} & \underline{50.13}  & 34.30          & 50.44          & 49.83          & 32.11          & 32.57          & 36.78          & 54.32          & 51.41          & 62.51          & 40.08          & 48.56           \\
DMNet (TGRS'23)\cite{bi2023not}                                   & 51.41          & 56.87          & 56.95          & 66.03          & 40.99          & 33.17          & 49.18          & 49.66          & 36.06          & 35.33          & 51.14          & \textbf{71.12} & 50.46          & \underline{67.01}  & 28.27          & 49.58           \\
\rowcolor[rgb]{0.941,0.941,0.941} \textbf{ViRefSAM (Ours)} & \underline{52.68}  & \textbf{64.32} & 60.12          & \underline{66.87}  & 45.61          & \underline{37.8}   & \textbf{54.14} & \textbf{57.38} & \textbf{38.11} & 36.27          & 60.74          & 69.51          & 60.48          & 58.62          & 33.2           & \textbf{53.06}  \\ 
\hline
\multicolumn{17}{c}{ResNet-101}                                                                                                                                                                                                                                                                                                    \\ 
\hline
ASGNet (CVPR'21)\cite{li2021adaptive}                                  & 42.81          & 50.90          & 57.23          & 34.69          & \underline{52.15}  & \underline{41.00}  & 52.03          & 35.95          & \underline{36.22}  & 27.16          & \underline{66.46}  & 63.57          & 45.04          & 21.20          & \underline{50.15}  & 45.10           \\
CyCTR (NeurIPS'21)\cite{zhang2021few}                                & 48.76          & 49.24          & 58.79          & 45.06          & \textbf{52.59} & \textbf{52.01} & 45.42          & 33.52          & 34.82          & 28.7           & \textbf{68.74} & 43.47          & \textbf{66.18} & 24.37          & \textbf{58.24} & 47.33           \\
PFENet (TPAMI'22)\cite{tian2022prior}                                 & 46.63          & 48.55          & 57.02          & 59.99          & 41.25          & 27.32          & 42.42          & 48.66          & 34.63          & 38.82          & 54.78          & 70.40          & 51.71          & 64.30          & 23.05          & 47.30           \\
NERTNet (CVPR'22)\cite{liu2022learning} & 50.41          & 48.21          & 52.63          & 61.77          & 38.61          & 29.17          & 37.44          & 53.25          & 34.79          & \underline{39.03}  & 41.05          & \underline{72.40}  & 59.08          & 61.30          & 22.31          & 46.76           \\
DCPNet (IJCV'23)\cite{lang2024few}                                  & 49.24          & 39.46          & 47.46          & 59.32          & 42.67          & 24.16          & 48.82          & 46.89          & 34.74          & \textbf{39.40} & 37.59          & 64.53          & 58.92          & 62.70          & 22.95          & 45.26           \\
SCCAN (ICCV'23)\cite{xu2023self}                                   & 48.70          & 55.04          & 57.39          & 57.34          & 45.08          & 30.00          & 44.29          & 52.87          & 31.90          & 34.73          & 42.46          & \textbf{78.50} & 61.31          & 61.54          & 23.44          & 48.31           \\
BAM (TPAMI'23)\cite{lang2023base}                                    & \underline{55.03}  & \textbf{63.48} & \underline{62.01}  & \underline{65.57}  & 49.38          & 25.81          & \textbf{55.12} & \textbf{58.04} & 30.82          & 38.47          & 58.56          & 64.01          & 52.51          & \textbf{65.60} & 12.04          & \underline{50.43}   \\
R2Net (TGRS'23)\cite{lang2024few}                                   & \textbf{56.33} & 61.36          & \textbf{68.92} & 60.11          & 41.97          & 34.86          & 45.39          & 51.31          & 28.47          & 38.74          & 52.25          & 67.48          & 48.73          & 60.36          & 13.43          & 48.65           \\
DMNet (TGRS'23)\cite{bi2023not}                                   & 51.98          & 56.31          & 56.98          & \textbf{67.13} & 37.65          & 32.97          & 48.21          & 49.23          & 34.25          & 35.54          & 45.89          & 70.04          & 57.86          & \underline{65.50}  & 28.56          & 49.21           \\
\rowcolor[rgb]{0.941,0.941,0.941} \textbf{ViRefSAM (Ours)} & 53.65          & \underline{62.87}  & 61.21          & 64.34          & 43.43          & 36.77          & \underline{54.95}  & \underline{56.24}  & \textbf{38.47} & 35.72          & 62.28          & 68.54          & \underline{61.33}  & 59.07          & 32.98          & \textbf{52.79}  \\ 
\hline
\multicolumn{17}{c}{DeiT-B/16}                                                                                                                                                                                                                                                                                                     \\ 
\hline
FPTrans
(NeurIPS'22)\cite{zhang2022feature}                            & 44.56          & 63.97          & 54.83          & 67.54          & 27.07          & 27.49          & 49.24          & 45.93          & \underline{37.66}  & 28.62          & \underline{45.40}  & \textbf{84.63} & 41.39          & \underline{66.75}  & 26.11          & 47.41           \\
AgMTR (IJCV'24)\cite{bi2024agmtr}                                   & \underline{49.22}  & \textbf{68.55} & \underline{58.99}  & \textbf{70.37} & \underline{40.41}  & \underline{30.73}  & \textbf{61.07} & \underline{50.55}  & 37.37          & \textbf{37.99} & 30.08          & \underline{83.14}  & \underline{56.73}  & \textbf{67.26} & \textbf{31.23} & \underline{51.58}   \\
\rowcolor[rgb]{0.941,0.941,0.941} \textbf{ViRefSAM (Ours)} & \textbf{52.35} & \underline{67.49}  & \textbf{60.21} & \underline{67.79}  & \textbf{42.81} & \textbf{34.40} & \underline{57.49}  & \textbf{58.21} & \textbf{40.53} & \underline{35.67}  & \textbf{48.73} & 79.55          & \textbf{57.48} & 64.62          & \underline{29.67}  & \textbf{53.13}  \\
\bottomrule
\end{tabular}
\begin{tablenotes}
\footnotesize
\item[] Note that the categories of C1-C15 are: ``ship", ``storage tank", ``baseball diamond", ``tennis court", ``basketball court", ``ground track field", ``bridge", ``large vehicle", ``small vehicle", ``helicopter", ``swimming pool", ``roundabout", ``soccer ball field", ``plane", ``harbor".
\end{tablenotes}
\end{threeparttable}
}
\end{table*}

\section{Experiments}\label{sec:Experiments}

To begin with, Sec.\ref{sec: Experimental Setup} introduces the experimental settings, including dataset descriptions, implementation details, and evaluation metrics. To validate the effectiveness of the proposed method, extensive comparisons with state-of-the-art methods are presented in Sec.\ref{sec: Comparison with State-of-the-arts}. Finally, Sec.\ref{sec: Ablation Study and Analysis} provides a detailed ablation study to further analyze the contributions of each component in ViRefSAM.

\subsection{Experimental Setup}\label{sec: Experimental Setup}
\subsubsection{Datasets}\label{sec:Datasets} 
Following the few-shot setting, we conduct comprehensive performance evaluations on three FSS benchmark datasets, including two RS datasets (iSAID-5$^i$~\cite{bi2023not} and LoveDA-2$^i$~\cite{bi2023not}) and one CV dataset (COCO-20$^i$~\cite{tian2022prior}).

\noindent \textbf{iSAID-5$^i$} is derived from the iSAID segmentation dataset~\cite{waqas2019isaid}, which contains 655,451 object instances across 15 categories in 2,806 high-resolution RS images. Following the DMNet setup~\cite{bi2023not}, the 15 categories are divided into three folds, each with 10 training classes and 5 testing classes. The training and testing classes are mutually exclusive within each fold to simulate the unseen-class scenario. For evaluation, 1,000 reference-target pairs are randomly sampled from each fold.

\noindent \textbf{LoveDA-2$^i$}, is built upon the LoveDA dataset~\cite{wang2021loveda}, consisting of 5,987 high-resolution images and 166,768 labeled objects across urban and rural regions. The dataset includes seven semantic categories (e.g., ``road'', ``water’’, ``barren'', ``forest’’). Following~\cite{bi2023not}, we exclude the background and split the remaining six classes into three folds, each with four classes for training and two for testing. All images are randomly cropped to 473$\times$473. During evaluation, 1,000 reference-target pairs are sampled from each fold.

\noindent \textbf{COCO-20$^i$}, is constructed from the MSCOCO dataset~\cite{lin2014microsoft}, which contains 80 object classes. Following the protocol in~\cite{lang2024few}, the classes are divided into four folds, each with 60 classes for training and 20 for testing. For each fold, 1,000 reference-target pairs are randomly sampled for validation.

\begin{table*}[t]
\setlength{\abovecaptionskip}{5pt}
\setlength{\belowcaptionskip}{10pt}
\caption{ Performance comparison on LoveDA-2$^i$ under the 1-shot and 5-shot settings, utilizing mIoU and FB-IoU (\%) as evaluation metrics. ``Mean” denotes the average mIoU across the three folds, and ``FB-IoU” represents the average foreground-background IoU over the same folds.} \label{tab:3}
\renewcommand\arraystretch{1.25}
\centering
\resizebox{1.0\linewidth}{!}{
\begin{tabular}{cr|ccccc|ccccc} 
\toprule
\multirow{2}{*}{Backbone}   & \multirow{2}{*}{Method}                                      & \multicolumn{5}{c|}{1-shot}                                                                                                                                                                                                                                            & \multicolumn{5}{c}{5-shot}                                                                                                                                                                                                                                              \\ 
\cline{3-12}
&                                                              & Fold-0                                             & Fold-1                                             & Fold-2                                             & Mean                                               & FB-IoU                                             & Fold-0                                             & Fold-1                                             & Fold-2                                             & Mean                                               & FB-IoU                                              \\ 
\hline
\multirow{9}{*}{ResNet-50}  & SCL (CVPR'21)\cite{zhang2021self}                                                & 15.14                                              & 20.45                                              & 25.00                                              & 20.20                                              & 24.60                                              & 14.25                                              & 21.09                                              & 23.65                                              & 19.66                                              & 24.69                                               \\
& ASGNet (CVPR'21)\cite{li2021adaptive}                                             & 15.91                                              & 20.21                                              & 22.33                                              & 19.48                                              & 36.39                                              & 18.38                                              & 26.29                                              & 36.34                                              & 27.00                                              & 39.59                                               \\
& CyCTR (NeurIPS'21)\cite{zhang2021few}                                           & 13.17                                              & 23.43                                              & 21.99                                              & 19.53                                              & 38.47                                              & 13.81                                              & 27.40                                              & 26.15                                              & 22.45                                              & 42.71                                               \\
& PFENet (TPAMI'22)\cite{tian2022prior}                                            & 17.13                                              & 22.20                                              & 26.49                                              & 21.94                                              & 33.48                                              & 15.83                                              & 25.73                                              & 24.74                                              & 22.10                                              & 34.77                                               \\
& NERTNet (CVPR'22)\cite{liu2022learning}                                            & 16.05                                              & 22.69                                              & 21.87                                              & 20.20                                              & 32.67                                              & 15.79                                              & 24.94                                              & 23.42                                              & 21.38                                              & 31.79                                               \\
& DCPNet (IJCV'23)\cite{lang2024few}                                             & 16.67                                              & 23.10                                              & 24.44                                              & 21.40                                              & 36.36                                              & 13.43                                              & 25.59                                              & 28.06                                              & 22.36                                              & 33.89                                               \\

& DMNet (TGRS'23)\cite{bi2023not}                                              & 19.29                                              & 25.52                                              & 31.53                                              & 25.45                                              & 43.40                                              & 24.62                                              & 33.80                                              & 33.12                                              & 30.51                                              & 50.93                                               \\
& MGANet (TGRS'24)\cite{MGANet}                                             & 20.55                                              & 27.51                                              & 33.30                                              & 27.12                                              & 44.91                                              & 26.53                                              & 35.99                                              & 36.69                                              & 33.07                                              & 50.68                                               \\
& {\cellcolor[rgb]{0.941,0.941,0.941}}\textbf{ViRefSAM (Ours)} & {\cellcolor[rgb]{0.941,0.941,0.941}}\textbf{22.09} & {\cellcolor[rgb]{0.941,0.941,0.941}}\textbf{28.94} & {\cellcolor[rgb]{0.941,0.941,0.941}}\textbf{34.77} & {\cellcolor[rgb]{0.941,0.941,0.941}}\textbf{28.60} & {\cellcolor[rgb]{0.941,0.941,0.941}}\textbf{45.51} & {\cellcolor[rgb]{0.941,0.941,0.941}}\textbf{28.68} & {\cellcolor[rgb]{0.941,0.941,0.941}}\textbf{36.25} & {\cellcolor[rgb]{0.941,0.941,0.941}}\textbf{37.48} & {\cellcolor[rgb]{0.941,0.941,0.941}}\textbf{34.14} & {\cellcolor[rgb]{0.941,0.941,0.941}}\textbf{51.80}  \\ 
\hline
\multirow{9}{*}{ResNet-101} & SCL (CVPR'21)\cite{zhang2021self}                                                & 15.62                                              & 17.87                                              & 25.54                                              & 19.68                                              & 31.03                                              & 14.96                                              & 20.26                                              & 24.02                                              & 19.75                                              & 31.57                                               \\
& ASGNet (CVPR'21)\cite{li2021adaptive}                                             & 14.65                                              & 19.90                                              & 25.43                                              & 19.99                                              & 30.89                                              & 18.00                                              & 25.43                                              & 36.00                                              & 26.48                                              & 39.26                                               \\
& CyCTR (NeurIPS'21)\cite{zhang2021few}                                           & 13.16                                              & 20.63                                              & 20.55                                              & 18.11                                              & 38.94                                              & 15.40                                              & 25.27                                              & 22.01                                              & 20.89                                              & 36.03                                               \\
& PFENet (TPAMI'22)\cite{tian2022prior}                                            & 15.83                                              & 25.73                                              & 24.74                                              & 22.10                                              & 35.65                                              & 15.62                                              & 24.37                                              & 26.64                                              & 22.21                                              & 36.08                                               \\
& NERTNet (CVPR'22)\cite{liu2022learning}                                            & 15.51                                              & 19.65                                              & 30.07                                              & 21.74                                              & 33.68                                              & 15.18                                              & 23.50                                              & 31.83                                              & 23.50                                              & 32.16                                               \\
& DCPNet (IJCV'23)\cite{lang2024few}                                             & 16.52                                              & 20.20                                              & 33.61                                              & 23.44                                              & 36.57                                              & 16.97                                              & 25.08                                              & 25.10                                              & 22.38                                              & 35.71                                               \\

& DMNet (TGRS'23)\cite{bi2023not}                                              & 21.18                                              & 24.57                                              & 28.74                                              & 24.83                                              & 47.47                                              & 24.65                                              & 33.97                                              & 36.89                                              & 31.84                                              & 53.38                                               \\
& MGANet (TGRS'24)\cite{MGANet}                                             & 22.97                                              & 27.79                                              & 28.14                                              & 26.30                                              & 46.83                                              & 26.29                                              & 34.88                                              & 36.84                                              & 32.67                                              & 52.88                                               \\
& {\cellcolor[rgb]{0.941,0.941,0.941}}\textbf{ViRefSAM (Ours)} & {\cellcolor[rgb]{0.941,0.941,0.941}}\textbf{23.53} & {\cellcolor[rgb]{0.941,0.941,0.941}}\textbf{29.82} & {\cellcolor[rgb]{0.941,0.941,0.941}}\textbf{31.62} & {\cellcolor[rgb]{0.941,0.941,0.941}}\textbf{28.32} & {\cellcolor[rgb]{0.941,0.941,0.941}}\textbf{46.12} & {\cellcolor[rgb]{0.941,0.941,0.941}}\textbf{27.45} & {\cellcolor[rgb]{0.941,0.941,0.941}}\textbf{35.74} & {\cellcolor[rgb]{0.941,0.941,0.941}}\textbf{36.66} & {\cellcolor[rgb]{0.941,0.941,0.941}}\textbf{33.28} & {\cellcolor[rgb]{0.941,0.941,0.941}}\textbf{53.45}  \\
\bottomrule
\end{tabular}}
\end{table*}

\subsubsection{Implementation Details}\label{sec:Implementation Details}
The proposed ViRefSAM framework is implemented utilizing the PyTorch library~\cite{paszke2019pytorch} and trained on an NVIDIA Tesla A100 GPU (40G). The model is optimized with the AdamW optimizer, using a batch size of 8 and an initial learning rate of 2e-4. A cosine learning rate decay schedule is applied throughout training. For the iSAID dataset, we train the model for 70 epochs; for LoveDA-2$^i$ and COCO-20$^i$, the total number of training epochs is set to 50. All input images are resized to 512$\times$512 to comply with SAM’s input requirements. To ensure fair comparison with previous works, we follow the same data augmentation strategies adopted in prior FSS literature.

For the SAM model, we adopt the SAM-Huge architecture. As for the visual contextual prompt encoder, we experiment with four backbone architectures, including ResNet-50, ResNet-101, ViT-B/16~\cite{dosovitskiy2020image}, and DeiT-B/16~\cite{touvron2021training}, all of which are pre-trained on ImageNet~\cite{deng2009imagenet}. This allows direct comparison with various FSS methods utilizing different encoder types. In our experiments, the number of background prototypes $N_b$ is set to 5, the number of learnable queries $N_q$ is set to 64, and the weight $\gamma$ for the query regularization loss is set to 0.2.

\subsubsection{Evaluation Metrics}\label{sec:Evaluation Metrics}
Following prior work in few-shot segmentation~\cite{lang2023base}, we adopt two commonly used evaluation metrics: mean Intersection-over-Union (mIoU) and Foreground-Background IoU (FB-IoU). Specifically, mIoU measures the average segmentation performance across all unseen classes and is defined as: $\mathrm{mIoU}=\frac{1}{n}\sum_{i=1}^{N_\mathrm{unseen}}\mathrm{IoU_i}$, where $N_{\mathrm{unseen}}$ is the number of unseen classes in each evaluation fold and $\mathrm{IoU}i$ denotes the IoU score for class $i$. FB-IoU provides a coarser-grained evaluation by averaging the segmentation performance over foreground and background regions, calculated as: $\mathrm{FB}$-$\mathrm{IoU} =\left ( \mathrm{IoU_{F}}+ \mathrm{IoU_{B}}  \right )/2$, where $\mathrm{IoU}_{\mathrm{F}}$ and $\mathrm{IoU}_{\mathrm{B}}$ represent the IoU scores for the foreground and background, respectively. Considering its fine-grained class-level granularity and broader applicability, mIoU is selected as the primary evaluation metric throughout our experiments.

\subsection{Comparison with State-of-the-arts}\label{sec: Comparison with State-of-the-arts}

\subsubsection{Quantitative Results}\label{sec:Quantitative Results}
In this section, we present quantitative evaluations on three few-shot segmentation (FSS) datasets. Specifically, we compare the proposed ViRefSAM with existing FSS and RS-FSS methods on two RS benchmarks, iSAID-5$^i$ and LoveDA-2$^i$. Furthermore, we assess the generalization capability of ViRefSAM beyond the RS domain by evaluating its performance alongside several foundation models on the COCO-20$^i$ dataset.

\noindent \textbf{iSAID-5$^i$.} 
As depicted in Table \ref{tab:1}, we report the 1-shot and 5-shot segmentation results of the proposed ViRefSAM on the iSAID-5$^i$ dataset, compared with several competitive methods. It is evident that ViRefSAM consistently achieves state-of-the-art performance under all backbone settings. For example, under the 1-shot setting with the ResNet-50 backbone, ViRefSAM attains 53.06\% mIoU, outperforming the previous best method MGANet~\cite{MGANet} (51.85\%). When adopting the DeiT-B/16 backbone, it further achieves 53.13\% mIoU, surpassing AgMTR~\cite{bi2024agmtr} by 1.55\%. Similar performance gains are observed across other backbones. Under the 5-shot setting, ViRefSAM also demonstrates superior results in both mIoU and FB-IoU metrics. Notably, in some cases, our method under the 1-shot setting even outperforms other methods under the 5-shot setting, such as R2Net~\cite{lang2023global} and DMNet~\cite{bi2023not}. These results highlight the strong segmentation capability and generalization of ViRefSAM, which can accurately identify class-specific objects in RS images under the guidance of reference images, even for unseen classes.

To further demonstrate the segmentation capability of ViRefSAM, Table \ref{tab:2} presents the class-wise IoU scores on all three folds of the iSAID-5$^i$ dataset. Specifically, classes C1–C5 correspond to the testing classes in Fold 0, C6–C10 to Fold 1, and C11–C15 to Fold 2. It can be observed that ViRefSAM ranks among the top three in most categories, consistently achieving competitive or superior performance across diverse object types. This highlights the model’s strong generalization ability in fine-grained RS segmentation tasks under limited supervision.

\noindent \textbf{LoveDA-2$^i$.} 
Table \ref{tab:3} reports the 1-shot and 5-shot segmentation results on the LoveDA-2$^i$ dataset. As shown, the overall mIoU scores are relatively lower compared to other benchmarks, primarily due to the inherent domain shift in LoveDA, which includes heterogeneous urban and rural landscapes. Moreover, the dataset focuses on land cover categories rather than well-defined foreground objects (e.g., ``plane"), posing greater challenges for model generalization. Despite these difficulties, ViRefSAM consistently achieves superior performance. With the ResNet-50 backbone, it surpasses competing methods by 1.48\% and 1.07\% mIoU under the 1-shot and 5-shot settings, respectively. Similar gains are observed with the ResNet-101 backbone. These results demonstrate the robustness and strong generalization ability of ViRefSAM in handling complex cross-domain RS segmentation scenarios.

\noindent \textbf{COCO-20$^i$.} 
To further evaluate the generalizability and applicability of the proposed method beyond remote sensing, we conducted experiments on the more commonly used COCO-20$^i$ dataset. Table \ref{tab:4} presents the 1-shot and 5-shot segmentation results. In addition to comparing ViRefSAM with state-of-the-art FSS methods, we also include recent approaches leveraging vision foundation models for few-shot segmentation, such as SegGPT~\cite{wang2023seggpt}, PerSAM~\cite{zhang2023personalize}, and LLaFS~\cite{zhu2024llafs}. ViRefSAM achieves a competitive mIoU of 54.13\% under the 1-shot setting, outperforming SAM-based methods including PerSAM (23.03\%), Matcher (52.73\%), and VRP-SAM (53.88\%), as well as LLaFS (53.95\%), which incorporates large language models. Notably, methods like Painter~\cite{wang2023images} and SegGPT~\cite{wang2023seggpt} are trained utilizing all COCO categories, while ViRefSAM achieves strong performance despite being evaluated on entirely unseen categories, highlighting its robust generalization capabilities under the few-shot setting.

\subsubsection{Qualitative Results}\label{sec:Qualitative Results}
To further demonstrate the superiority of the proposed ViRefSAM, we visualize representative segmentation results across three FSS datasets. As depicted in Fig.\ref{fig:7}, ViRefSAM achieves high-quality segmentation on iSAID-5$^i$, clearly outperforming existing methods. Compared to current FSS methods such as R2Net~\cite{lang2023global}, our method more accurately localizes and delineates the target objects while suppressing irrelevant regions. For instance, in the second column, ViRefSAM effectively segments densely arranged ``ships", while R2Net fails to capture precise object boundaries. Similarly, in the third column, our method accurately segments the roundabout structure, while R2Net incorrectly activates irrelevant categories such as “parking lots”. These improvements are primarily attributed to the proposed visual contextual prompt encoder, which effectively leverages reference images to activate SAM’s prompt-driven generalization capabilities. Moreover, the dynamic target alignment adapter further enhances SAM’s focus on task-relevant semantics, resulting in more precise segmentation. Fig.\ref{fig:8} and Fig.\ref{fig:9} show qualitative results on LoveDA-2$^i$ and COCO-20$^i$, respectively. ViRefSAM consistently delivers robust performance across diverse and challenging domains, confirming its strong generalization ability and applicability to both cross-domain RS tasks and natural image scenarios.

\subsection{Ablation Study and Analysis}\label{sec: Ablation Study and Analysis}
We conduct a series of ablation studies to further investigate and understand the effectiveness of the proposed method. Unless otherwise specified, all experiments in this section are performed on the iSAID-5$^i$ dataset under the 1-shot setting using the ResNet-50 backbone.

\subsubsection{Performance under Various Reference Annotation Schemes}
To further alleviate the burden of dense annotations in real-world scenarios, this section explores the impact of different annotation schemes for reference images on segmentation performance, including more economical alternatives such as points, bounding boxes, and scribbles. Following the setup in~\cite{wang2019panet,sun2024vrp}, these annotations are randomly generated from the original dense masks of the reference images. Table \ref{tab:5} and Fig.\ref{fig:10} present the quantitative and qualitative results under the 1-shot setting on three FSS benchmark datasets.

\begin{table*}[t]
\setlength{\abovecaptionskip}{5pt}
\setlength{\belowcaptionskip}{10pt}
\caption{ Performance comparison on COCO-20$^i$ under the 1-shot and 5-shot settings, utilizing mIoU (\%) as evaluation metrics. ``Mean” denotes the average mIoU across the four folds. \textbf{Bold} and \underline{underlined} values indicate the best and second-best performance, respectively. Notably, Painter~\cite{wang2023images} and SegGPT~\cite{wang2023seggpt} were trained with access to all COCO categories, while the other methods maintain evaluations on unseen classes. } \label{tab:4}
\renewcommand\arraystretch{1.25}
\centering
\resizebox{0.7\linewidth}{!}{
\begin{tabular}{c|r|cccc|c} 
\toprule
\multicolumn{2}{c|}{Method}                                                                                                            & Fold-0                                    & Fold-1                                    & Fold-2                                    & Fold-3                                    & Mean                 \\ 
\hline
\multirow{3}{*}{\begin{tabular}[c]{@{}c@{}}Specialized\\FSS Model\end{tabular}} & DBMNet (TIP'24)\cite{chen2024dual}                                       & 41.80                                     & 45.60                                     & 43.20                                     & 41.30                                     & 42.98                                      \\
& HMNet (NeurIPS'24)\cite{xu2024hybrid}                                    & 45.50                                     & \underline{58.70}                             & 52.90                                     & 51.40                                     & 52.13                                      \\
& AgMTR (IJCV'24)\cite{bi2024agmtr}                                       & 44.01                                     & 55.51                                     & 49.90                                     & 46.61                                     & 49.01                                      \\ 
\hline
\multirow{9}{*}{\begin{tabular}[c]{@{}c@{}}Foundation\\Model\end{tabular}}     & \textcolor[rgb]{0.502,0.502,0.502}{Painter (CVPR'23)\cite{wang2023images}} & \textcolor[rgb]{0.502,0.502,0.502}{31.20} & \textcolor[rgb]{0.502,0.502,0.502}{35.30} & \textcolor[rgb]{0.502,0.502,0.502}{33.50} & \textcolor[rgb]{0.502,0.502,0.502}{32.40} & \textcolor[rgb]{0.502,0.502,0.502}{33.10}  \\
& \textcolor[rgb]{0.502,0.502,0.502}{SegGPT (ICCV'23)\cite{wang2023seggpt}}  & \textcolor[rgb]{0.502,0.502,0.502}{56.30} & \textcolor[rgb]{0.502,0.502,0.502}{57.40} & \textcolor[rgb]{0.502,0.502,0.502}{58.90} & \textcolor[rgb]{0.502,0.502,0.502}{51.70} & \textcolor[rgb]{0.502,0.502,0.502}{56.08}  \\
& PerSAM (ICLR'24)\cite{zhang2023personalize}                                      & 23.10                                     & 23.60                                     & 22.00                                     & 23.40                                     & 23.03                                      \\
& PerSAM-F (ICLR'24)\cite{zhang2023personalize}                                    & 22.30                                     & 24.00                                     & 23.40                                     & 24.10                                     & 23.45                                      \\
& Matcher (ICLR'24)\cite{li2024matching}                                     & \textbf{52.70}                            & 53.50                                     & 52.60                                     & \underline{52.10}                                     & 52.73                                      \\
& VRP-SAM (CVPR'24)\cite{sun2024vrp}                                     & 48.10                                     & 55.80                                     & \textbf{60.00}                            & 51.60                                     & 53.88                                      \\
& PAT (TPAMI'24)\cite{bi2024prompt}                                        & 40.62                                     & 51.94                                     & 48.99                                     & 50.66                                     & 48.05                                      \\
& LLaFS (CVPR'24)\cite{zhu2024llafs}                                       & 47.50                                     & \textbf{58.80}                            & 56.20                                     & \textbf{53.30}                            & \underline{53.95}                              \\ 
\cline{2-7}
& {\cellcolor[rgb]{0.941,0.941,0.941}}\textbf{ViRefSAM (Ours)} &{\cellcolor[rgb]{0.941,0.941,0.941}} \underline{49.41}                             & {\cellcolor[rgb]{0.941,0.941,0.941}}56.17                                     &{\cellcolor[rgb]{0.941,0.941,0.941}} \underline{59.19}                             &{\cellcolor[rgb]{0.941,0.941,0.941}} 51.74                             & {\cellcolor[rgb]{0.941,0.941,0.941}}\textbf{54.13}                             \\
\bottomrule
\end{tabular}}
\end{table*}

\begin{table*}[t]
\setlength{\abovecaptionskip}{5pt}
\setlength{\belowcaptionskip}{10pt}
\caption{Ablation study on the impact of different reference annotation schemes across three FSS datasets under the 1-shot setting. The mIoU (\%) is utilized as the evaluation metric. \textbf{Bold} and \underline{underlined} values indicate the best and second-best performance, respectively.} \label{tab:5}
\renewcommand\arraystretch{1.25}
\centering
\resizebox{1.0\linewidth}{!}{
\begin{tabular}{c|cccc|cccc|lllll} 
\toprule
\multirow{2}{*}{Annotation Scheme} & \multicolumn{4}{c|}{iSAID-5$^i$}                                        & \multicolumn{4}{c|}{LoveDA-2$^i$}                                       & \multicolumn{5}{c}{COCO-20$^i$}                                                            \\ 
\cline{2-14}
                            & Fold-0         & Fold-1         & Fold-2         & Mean           & Fold-0         & Fold-1         & Fold-2         & Mean           & Fold-0         & Fold-1         & Fold-2         & Fold-3         & Mean            \\ 
\hline
Point                       & 45.77          & 38.69          & 48.26          & 44.24          & 11.58          & 16.59          & 17.37          & 15.18          & 32.40          & 42.63          & 44.18          & 39.98          & 39.80           \\
Scribble                    & \underline{56.25}  & \underline{41.75}  & \underline{52.33}  & \underline{50.11}  & \underline{18.29}  & 22.19          & \underline{25.45}  & \underline{21.98}  & 44.75          & \underline{51.58}  & 53.63          & 47.30          & 49.32           \\
Box                         & 55.30          & 40.18          & 51.42          & 48.97          & 17.43          & \underline{22.22}  & 24.91          & 21.52          & \underline{45.21}  & 50.62          & \underline{54.49}  & \underline{48.13}  & \underline{49.61}   \\
Mask                        & \textbf{57.92} & \textbf{44.74} & \textbf{56.51} & \textbf{53.06} & \textbf{22.09} & \textbf{28.94} & \textbf{34.77} & \textbf{28.60} & \textbf{49.41} & \textbf{56.17} & \textbf{59.19} & \textbf{51.74} & \textbf{54.13}  \\
\bottomrule
\end{tabular}}
\end{table*}

\begin{figure}[t]
\setlength{\abovecaptionskip}{1pt} \setlength{\belowcaptionskip}{1pt}
\centering
\includegraphics[width=1.0\linewidth]{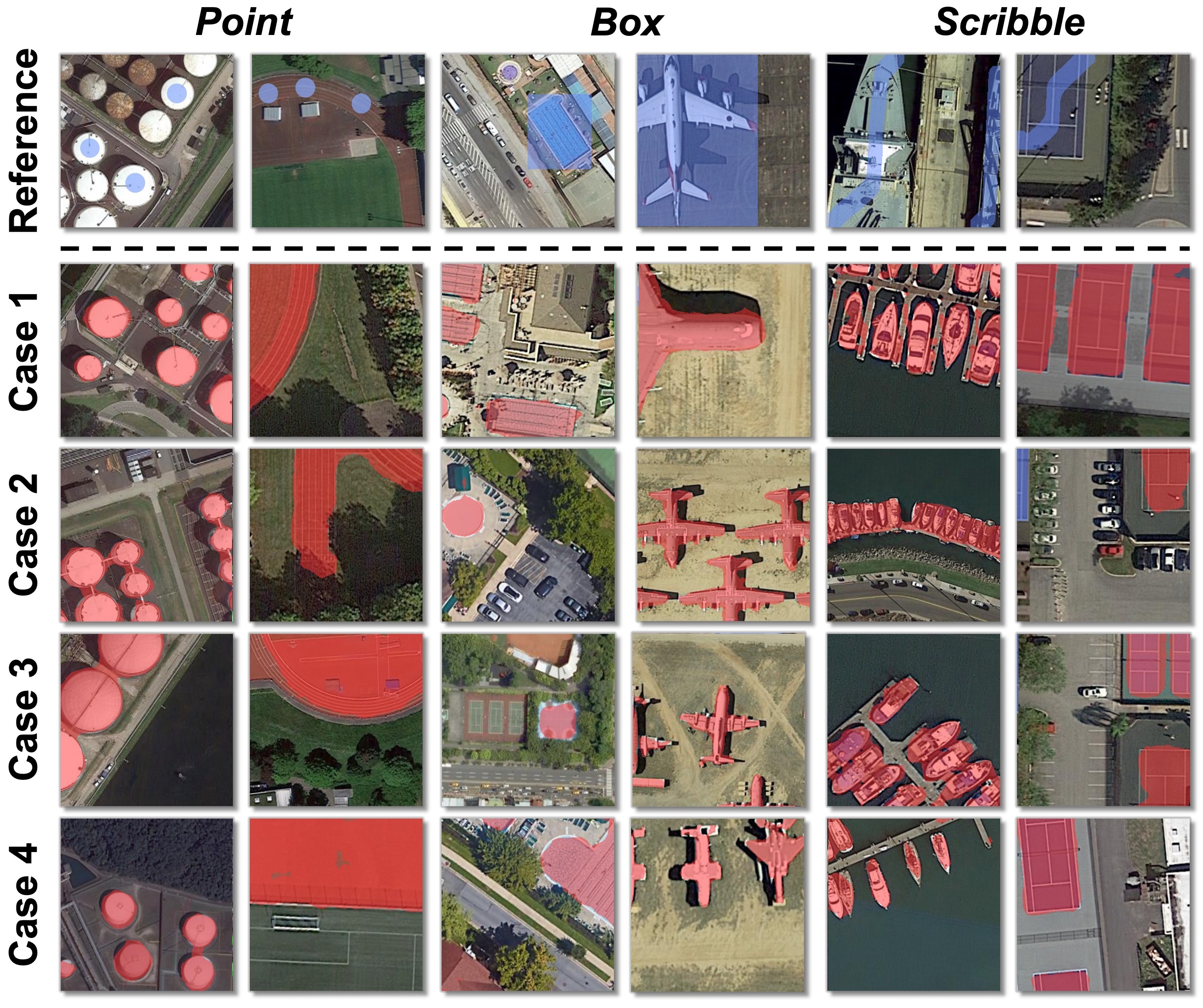}
\caption{\textbf{Qualitative visualization of different reference annotation schemes on the iSAID-5$^i$ benchmark under the 1-shot setting.} From top to bottom, reference images with ground truth annotations and four segmentation cases are provided.}
\label{fig:10}
\end{figure}

\begin{table}[t]
\setlength{\abovecaptionskip}{5pt}
\setlength{\belowcaptionskip}{10pt}
\caption{Ablation study of different components in ViRefSAM under the 1-shot setting. ``VCP Enc." refers to the Visual Contextual Prompt Encoder, and ``DTA Adap." denotes the Dynamic Target Alignment Adapter. ``Sample Prompts" indicates the types of prompts (e.g., points, boxes, or their combination) that are fed into the SAM's mask decoder for segmentation.
} \label{tab:6}
\renewcommand\arraystretch{1.25}
\centering
\resizebox{1.0\linewidth}{!}{
\begin{threeparttable}
\begin{tabular}{c|c|c|c} 
\toprule
\#                   & Methods                                                                           & Sample Prompts & mIoU(\%)            \\ 
\hline
\multirow{3}{*}{(a)} & \renewcommand\arraystretch{1.1}\multirow{3}{*}{ \begin{tabular}[c]{@{}c@{}}Reused Prompt-SAM \\(Fig.\ref{fig:1}(b))\end{tabular}}                                         & Point~                                & 45.24                                               \\
                     &                                                                                  & Box                                   & 43.80                                               \\
                     &                                                                                  & Point+Box                             & 45.38                                               \\ 
\hline
\multirow{3}{*}{(b)} & \renewcommand\arraystretch{1.1}\multirow{3}{*}{ \begin{tabular}[c]{@{}c@{}}SAM+PFENet \\(Baseline in Fig.\ref{fig:11})\end{tabular}} & Point~         & 48.12           \\
                     &                                                                                  & Box            & 47.95           \\
                     &                                                                                  & Point+Box      & 48.47           \\ 
\hline
(c)                  & SAM+VCP Enc.                                                                     & -              & \underline{51.61}           \\
(d)                  & SAM+PFENet+DTA Adap.$^\dagger$                                                             & Point+Box      & 51.32   \\
(e)                  & {\cellcolor[rgb]{0.941,0.941,0.941}}SAM+VCP Enc. +DTA Adap.                                                          & {\cellcolor[rgb]{0.941,0.941,0.941}}-              & {\cellcolor[rgb]{0.941,0.941,0.941}}\textbf{53.06}  \\
\bottomrule
\end{tabular}
\begin{tablenotes}
\footnotesize
\item[] Note $^\dagger$: In this variant, PFENet is integrated in place of the proposed VCP Encoder. Accordingly, the DTA adapter is reconfigured to fuse target features with reference prototypes and PFENet’s predicted prior mask, ensuring compatibility within the modified framework.
\end{tablenotes}
\end{threeparttable}

}
\end{table}
It can be observed that our ViRefSAM achieves mIoU scores of 50.11\%, 21.98\%, and 49.32\% on the three datasets using scribble annotations, which are even comparable to some FSS methods that rely on dense masks, such as PFENet~\cite{tian2022prior}. This demonstrates ViRefSAM’s strong tolerance to the quality of reference annotations. Moreover, we notice that the performance utilizing bounding box annotations is lower than that of scribble annotations. This may be attributed to the significant background noise introduced by bounding boxes, which hinders ViRefSAM’s ability to extract meaningful semantics from the reference image. Based on these observations, we conclude that scribble annotations offer a favorable trade-off between annotation cost and segmentation performance.

\subsubsection{Component Analysis}

This section investigates the performance contributions of different components in ViRefSAM through ablation studies. Specifically, we compare three variants: Reused Prompt-SAM (RP-SAM), the PFENet-based baseline, and our ViRefSAM.

We first introduce RP-SAM, a straightforward prompt reuse scheme commonly adopted in practical settings. As illustrated in Fig.\ref{fig:1}(b), RP-SAM directly reuses the prompts (e.g., points or bounding boxes) annotated on the reference image. These prompts are encoded by SAM’s prompt encoder and then applied to the target image through the SAM mask decoder to generate the segmentation result.

\begin{figure}[t]
\setlength{\abovecaptionskip}{2pt}
\centering
\includegraphics[width=0.85\linewidth]{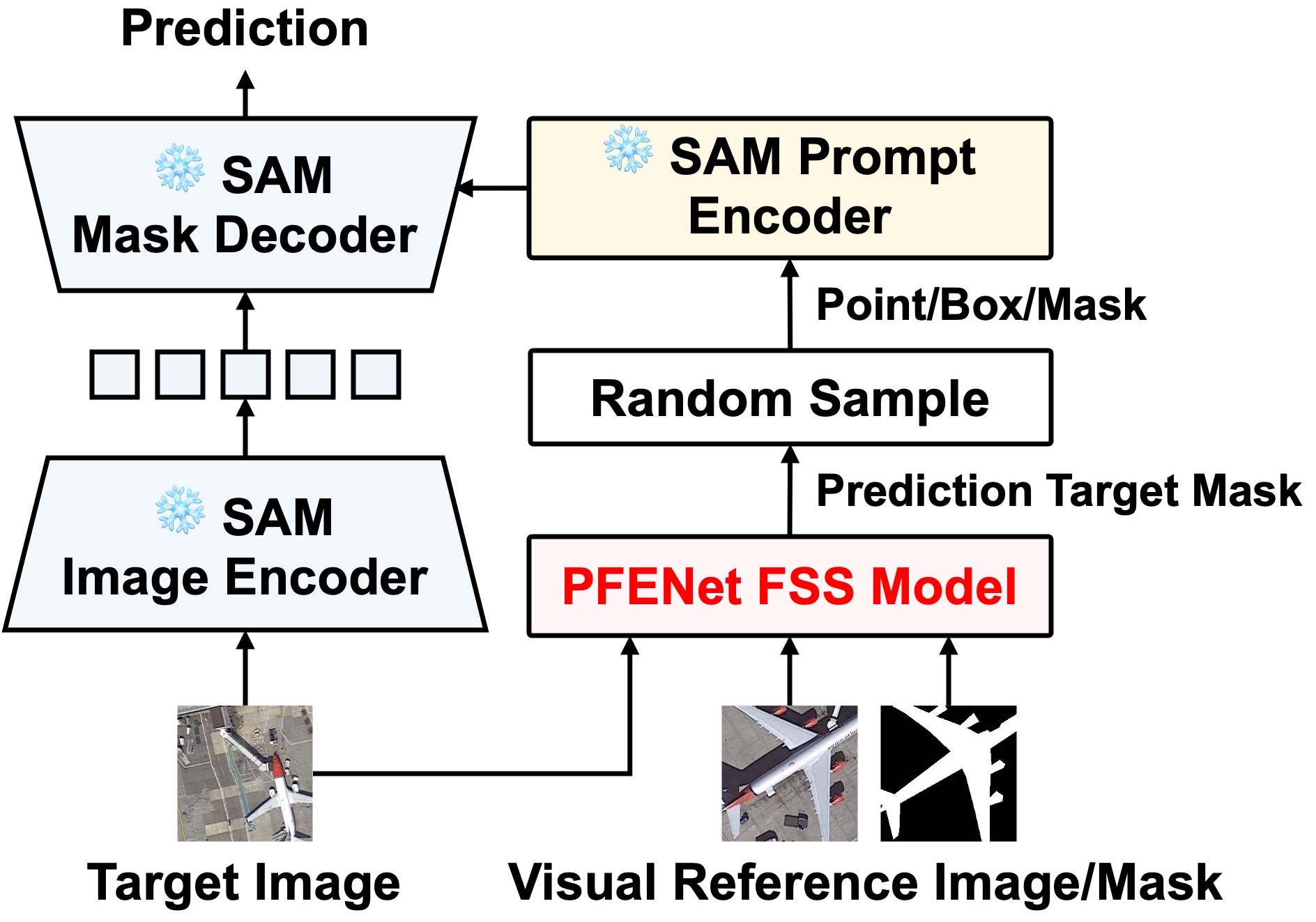}
\caption{\textbf{Baseline architecture of proposed ViRefSAM}. It combines SAM with PFENet, where PFENet generates pseudo masks from reference images. Various prompt types (e.g., points, boxes, and masks) are randomly sampled from these predicted masks and passed through SAM’s prompt encoder to generate prompt embeddings, which are then utilized by the SAM mask decoder to produce the final segmentation of the target image.}
\label{fig:11}
\end{figure}

Next, we define the baseline, which integrates SAM with the existing FSS method PFENet~\cite{tian2022prior}. As illustrated in Fig.\ref{fig:11}, given the target image and the reference image, both are input into PFENet to generate a pseudo mask for the target image. Various prompt types (e.g., point, box, and mask) are then randomly sampled from this predicted mask and encoded by SAM’s prompt encoder. The resulting prompt embeddings are passed to the SAM mask decoder to produce the final segmentation result.

\begin{figure}[t]
\setlength{\abovecaptionskip}{1pt} \setlength{\belowcaptionskip}{1pt}
\centering
\includegraphics[width=1.0\linewidth]{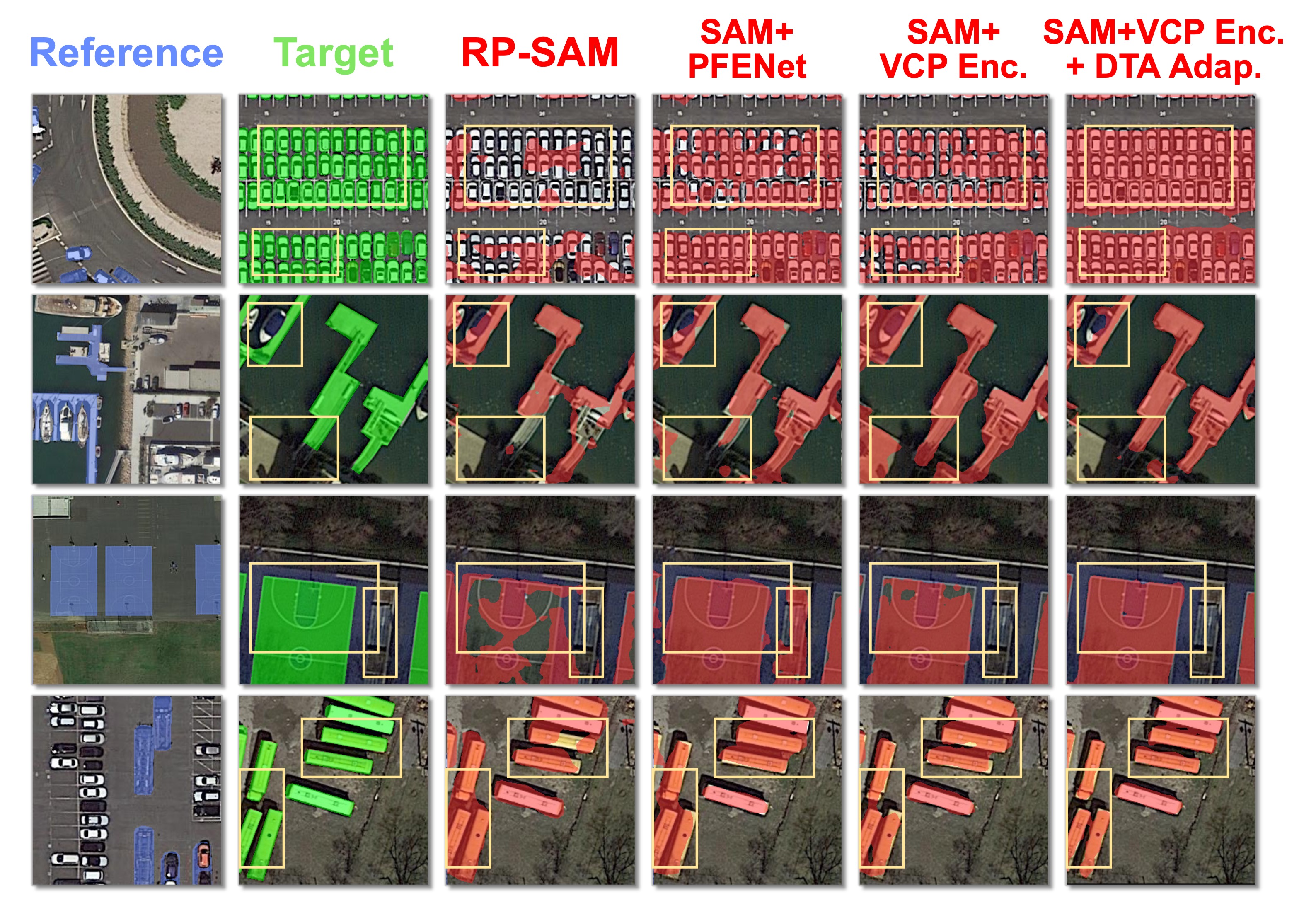}
\caption{\textbf{Qualitative visualization of ViRefSAM variants.} Each column, from left to right, shows: reference images with ground truth (GT), target image with GT, segmentation results from Reused Prompt-SAM (RP-SAM), SAM+PFENet, SAM+VCP Encoder, and SAM+VCP Encoder+DTA Adapter (i.e., the full ViRefSAM).}
\label{fig:12}
\end{figure}

Table \ref{tab:6} summarizes the quantitative results of different ViRefSAM variants under the 1-shot setting on iSAID-5$^i$. Overall, we observe the following: 
(i) Compared to Reused Prompt-SAM, which directly transfers prompts from the reference image to target images without adaptation, our ViRefSAM achieves significantly better performance. Specifically, the full version of ViRefSAM improves mIoU by 7.68\% over Reused Prompt-SAM, indicating that static prompt reuse is insufficient to handle the large semantic variations in RS scenes, and emphasizing the necessity of adaptive prompt generation guided by reference semantics. (ii) By incorporating the Visual Contextual Prompt (VCP) Encoder, our model achieves a 3.14\% mIoU improvement over the baseline. This clearly demonstrates the effectiveness of the VCP encoder and its better compatibility with the SAM architecture. (iii) Introducing the Dynamic Target Alignment (DTA) Adapter further improves performance by an additional 1.45\% mIoU, highlighting its ability to inject task-specific semantic clues into SAM’s image encoder and guide the model to focus on target-specific regions in RS images while maintaining generalization. (iv) As presented in Fig.~\ref{fig:12}, we also visualize segmentation results under different variants. With progressive integration of each component, the results show increasingly precise segmentation with better object focus and fewer false activations, qualitatively validating the effectiveness and necessity of the proposed components.

\begin{table}[t]
\setlength{\abovecaptionskip}{5pt}
\setlength{\belowcaptionskip}{10pt}
\caption{Ablation study on different variants of the proposed visual contextual encoder under the 1-shot setting. Both visual contextual interaction and object-aware prompt generation improve SAM’s segmentation performance.} \label{tab:7}
\renewcommand\arraystretch{1.25}
\centering
\resizebox{1.0\linewidth}{!}{
\begin{tabular}{c|cc|c} 
\toprule
\multirow{3}{*}{\#} & \multicolumn{2}{c|}{Visual Contextual Prompt Encoder}                                                                                           & \multirow{3}{*}{mIoU(\%)}  \\ 
\cline{2-3}
                    &\renewcommand\arraystretch{1.1} \begin{tabular}[c]{@{}c@{}}Visual Contextual\\ Interaction\end{tabular} & \renewcommand\arraystretch{1.1}\begin{tabular}[c]{@{}c@{}}Object-aware\\Prompt Generation\end{tabular} &                        \\ 
\hline
(a)                 & \textcolor[rgb]{0.702,0.702,0.702}{\ding{55}}                                 & \ding{51}                                                                     & 51.94                  \\
(b)                 & \ding{51}                                                                     & \textcolor[rgb]{0.702,0.702,0.702}{\ding{55}}                                 & \underline{52.43}          \\
(c)                 & {\cellcolor[rgb]{0.941,0.941,0.941}}\ding{51}                                                                     & {\cellcolor[rgb]{0.941,0.941,0.941}}\ding{51}                                                                     & {\cellcolor[rgb]{0.941,0.941,0.941}}\textbf{53.06}         \\
\bottomrule
\end{tabular}}
\end{table}

\begin{figure}[t]
\setlength{\abovecaptionskip}{2pt}
\centering
\includegraphics[width=0.95\linewidth]{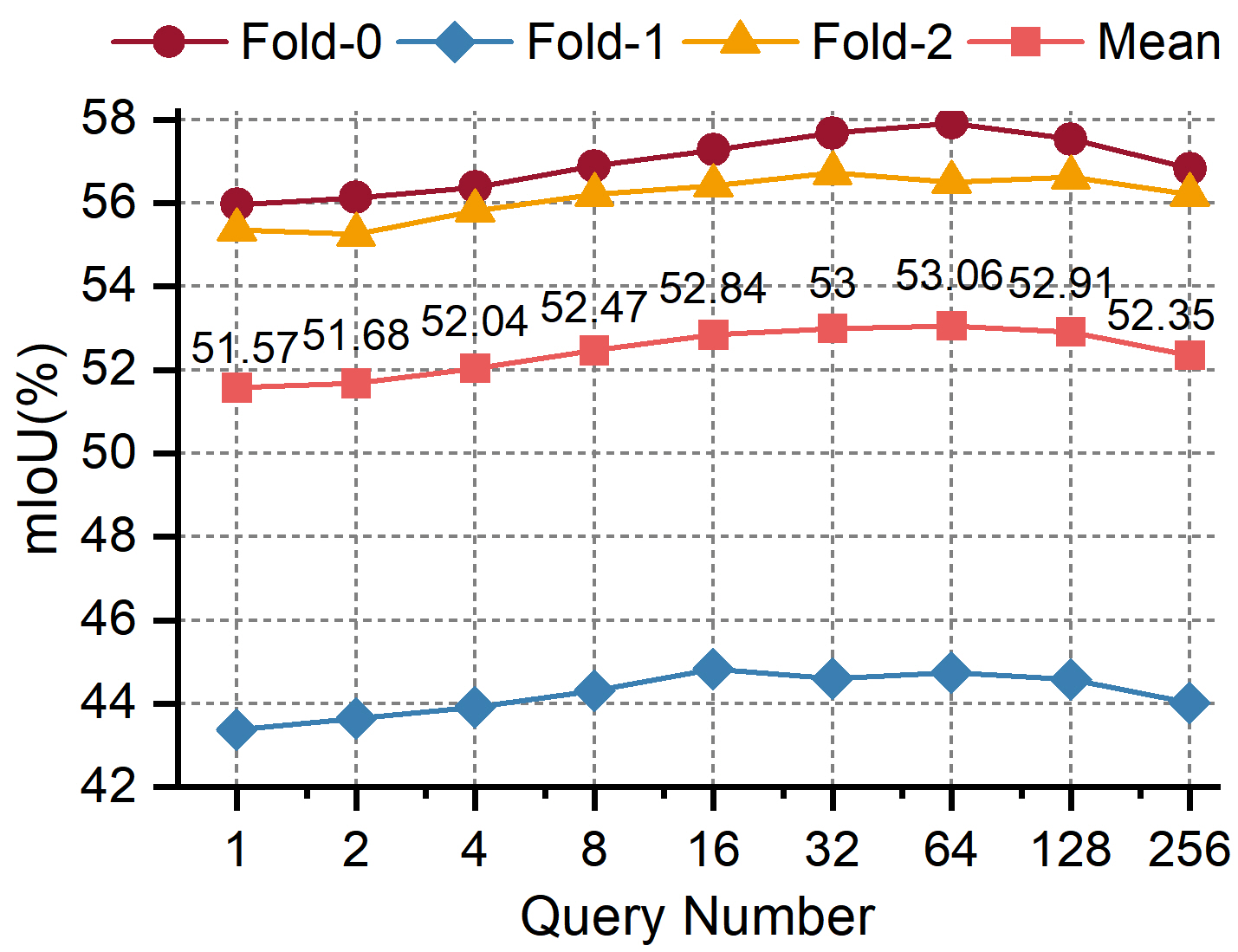}
\caption{\textbf{Ablation study on the number of learnable queries (i.e., $N_q$) in the object-aware prompt generation from VCP encoder}. The best segmentation performance is achieved at $N_q = 64$.}
\label{fig:13}
\end{figure}

\subsubsection{Ablation study of the Visual Contextual Prompt Encoder}

We first investigate the effect of different variants of the visual contextual encoder on segmentation performance. As presented in Table~\ref{tab:7}, by comparing case (c) with case (a), and case (c) with case (b), we observe that both the visual context interaction and the object-aware prompt generation effectively enhance the segmentation performance of SAM, yielding mIoU improvements of 1.12\% and 0.63\%, respectively. These results validate the effectiveness of the proposed components: the former captures class-specific clues from both reference and target images, while the latter generates object-aware embeddings aligned with the mask decoder. 

Next, we analyze the impact of the number of learnable queries (i.e., $N_q$) in the object-aware prompt generation, as illustrated in Fig.\ref{fig:13}. As $N_q$ increases from 1 to 64, the mIoU scores across the three-fold evaluation show a consistent upward trend, reaching a peak average mIoU of 53.06\% at $N_q = 64$. However, further increasing $N_q$ does not lead to additional performance gains and may even introduce redundancy or noise. Therefore, we set $N_q = 64$ in our experiments to balance segmentation accuracy and model efficiency.

\begin{table}[t]
\setlength{\abovecaptionskip}{5pt}
\setlength{\belowcaptionskip}{10pt}
\caption{Ablation study on different fine-tuning strategies for the SAM image encoder, including parameter freezing, full-parameter tuning, original adapter tuning, and the proposed Dynamic Target Alignment (DTA) Adapter tuning. $\Delta$ indicates the mIoU improvement over the parameter freezing baseline.} \label{tab:8}
\renewcommand\arraystretch{1.25}
\centering
\resizebox{1.0\linewidth}{!}{
\begin{tabular}{c|c|cc} 
\toprule
\#  & Fine-tuning Strategy               & mIoU(\%) & $\Delta$        \\ 
\hline
(a) & Freezing                          & 51.61    & -      \\
(b) & Full-parameter Tuning             & 47.19    & -4.42  \\
(c) & Adapter Tuning                    & 51.25    & -0.36  \\
\hline
(d) & DTA Adapter w/o Position Encoding & \underline{52.23}    & 0.62   \\
(e) & {\cellcolor[rgb]{0.941,0.941,0.941}} DTA Adapter Tuning                      & {\cellcolor[rgb]{0.941,0.941,0.941}} \textbf{53.06}    & {\cellcolor[rgb]{0.941,0.941,0.941}}1.45   \\
\bottomrule
\end{tabular}}
\end{table}

\begin{figure}[t]
\setlength{\abovecaptionskip}{1pt} \setlength{\belowcaptionskip}{1pt}
\centering
\includegraphics[width=1.0\linewidth]{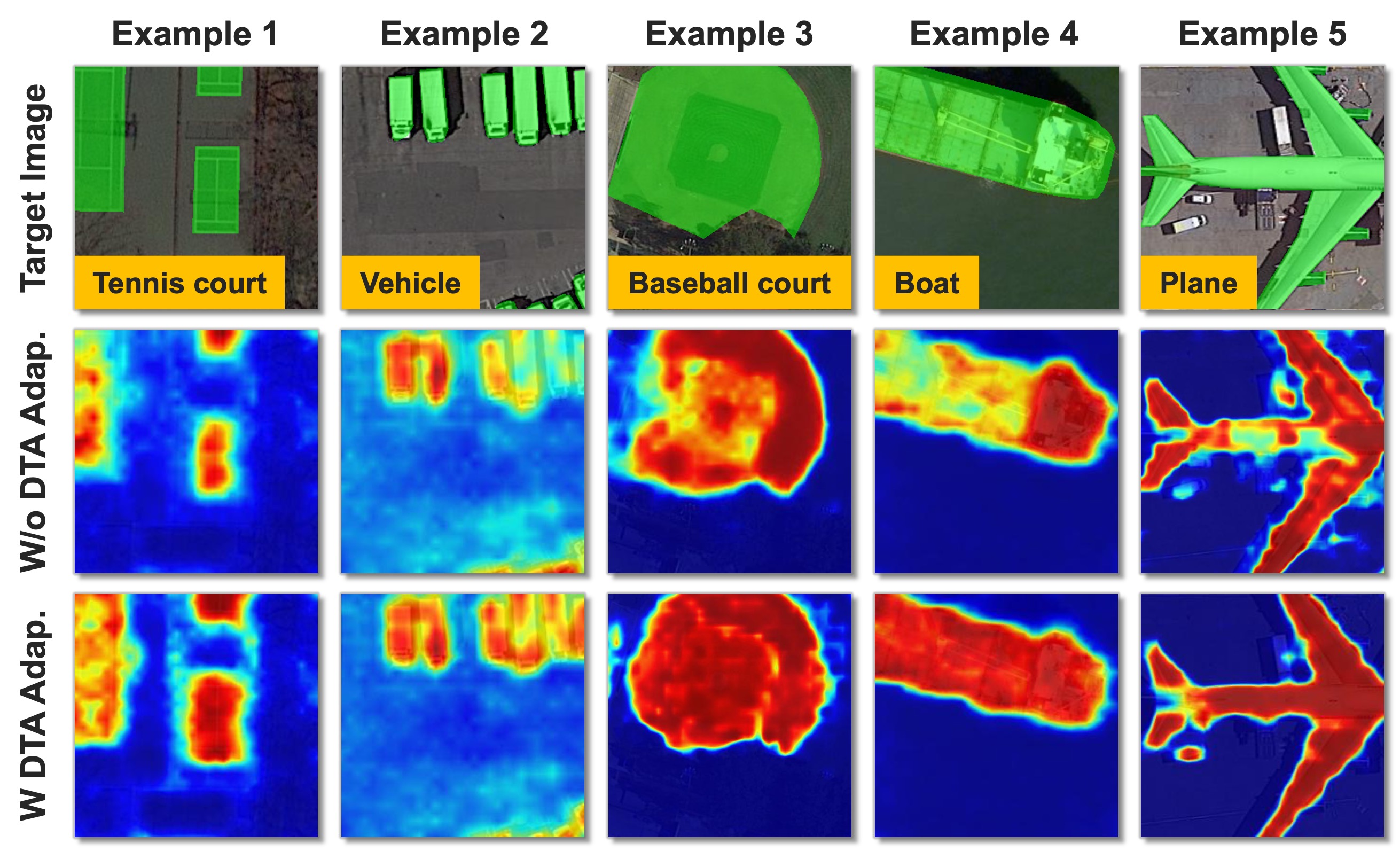}
\caption{\textbf{Qualitative visualization of the Dynamic Target Alignment (DTA) Adapter.} Five segmentation examples illustrate that the DTA adapter effectively guides SAM to focus on class-specific semantic regions in the current task, leading to more accurate object delineation.}
\label{fig:14}
\end{figure}

\begin{table}[t]
\setlength{\abovecaptionskip}{5pt}
\setlength{\belowcaptionskip}{10pt}
\caption{Ablation study of loss function in ViRefSAM under the 1-shot setting. The results demonstrate the contributions of BCE loss, Dice loss, and the regularization loss in improving segmentation performance and prompt diversity.} \label{tab:9}
\renewcommand\arraystretch{1.25}
\centering
\resizebox{1.0\linewidth}{!}{
\begin{tabular}{c|ccc|c} 
\toprule
\#  & BCE Loss & Dice Loss & Regularization & mIoU(\%)  \\ 
\hline
(a) &  \ding{51}     & \textcolor[rgb]{0.702,0.702,0.702}{\ding{55}}           & \textcolor[rgb]{0.702,0.702,0.702}{\ding{55}}                & 52.18     \\
(b) & \textcolor[rgb]{0.702,0.702,0.702}{\ding{55}}          &  \ding{51}      & \textcolor[rgb]{0.702,0.702,0.702}{\ding{55}}                & 52.25     \\
(c) &  \ding{51}     &  \ding{51}      & \textcolor[rgb]{0.702,0.702,0.702}{\ding{55}}                & \underline{52.61}     \\ 
(d) &  {\cellcolor[rgb]{0.941,0.941,0.941}}\ding{51}     &  {\cellcolor[rgb]{0.941,0.941,0.941}}\ding{51}      &  {\cellcolor[rgb]{0.941,0.941,0.941}}\ding{51}           & {\cellcolor[rgb]{0.941,0.941,0.941}}\textbf{53.06}     \\
\bottomrule
\end{tabular}}
\end{table}

\begin{table}[t]
\setlength{\abovecaptionskip}{5pt}
\setlength{\belowcaptionskip}{10pt}
\caption{Ablation study of different SAM architecture variants in ViRefSAM, showing ViT-Huge achieves the highest segmentation accuracy while ViT-Base offers the fastest inference speed.} \label{tab:10}
\renewcommand\arraystretch{1.25}
\centering
\resizebox{1.0\linewidth}{!}{
\begin{tabular}{ccccc} 
\toprule
\#                   & Encoder   & mIoU  & FB-IoU & Inference seed (FPS)  \\ 
\hline
\multirow{3}{*}{SAM} & ViT-Base  & 49.42 & 63.79  & 29.51                 \\
                     & ViT-Large & 51.89 & 66.50  & 22.68                 \\
                     & ViT-Huge  & 53.06 & 67.47  & 17.32                 \\
\bottomrule
\end{tabular}}
\end{table}

\subsubsection{Ablation study of the Dynamic Target Alignment Adapter}

In this section, we discuss the impact of different fine-tuning strategies applied to the SAM image encoder on segmentation performance, including parameter freezing, full-parameter tuning, original adapter tuning, and the proposed Dynamic Target Alignment (DTA) Adapter tuning. The results are summarized in Table~\ref{tab:8}. The key observations are as follows: (i) Fully fine-tuning the SAM image encoder leads to a significant performance degradation of 4.42\% mIoU. This may be attributed to the overfitting of the encoder to the known classes in the training set, thereby limiting its generalization ability to unseen classes. (ii) The original adapter tuning does not result in noticeable performance improvement, indicating that blindly inserting learnable parameters into the SAM encoder is insufficient for capturing transferable RS knowledge for unseen object segmentation. (iii) The proposed DTA adapter improves segmentation performance by 1.45\% mIoU. It injects class-specific semantic information from the current task into the SAM image features, enabling the model to dynamically focus on task-relevant objects. This facilitates effective knowledge adaptation in RS scenarios while preserving generalization ability. The qualitative results in Fig.\ref{fig:14} further support this finding. (iv) Additionally, introducing position encoding into the DTA adapter promotes better alignment between reference prototypes and target features, which further helps the model distinguish foreground and background regions effectively.

\subsubsection{Ablation study of Loss Function}

This section analyzes the impact of loss function design in ViRefSAM on segmentation performance. Table~\ref{tab:9} reports the experimental results on the iSAID-5$^i$ dataset under the 1-shot setting. The key observations are summarized as follows: 
(i) A comparison between case (a)-case (c) shows that both the binary cross-entropy (BCE) loss and Dice loss contribute positively to segmentation performance. Specifically, the BCE loss facilitates pixel-wise classification, while the Dice loss emphasizes region-level consistency. When combined, they provide complementary advantages and lead to further performance improvements.
(ii) Introducing a regularization loss (Eq.(\ref{equation:5})) in the object-aware prompt generation encourages the generated embeddings to be more semantically diverse and discriminative, effectively reducing semantic redundancy.

\subsubsection{Ablation study of SAM Structure Variants}
Table~\ref{tab:10} presents the impact of different SAM architectures on segmentation performance. It can be observed that using the ViT-Huge encoder achieves the highest segmentation accuracy, followed by ViT-Large, indicating that models with larger parameter counts tend to have stronger generalization ability and feature representation capacity. Additionally, the ViT-Base model offers the fastest inference speed, demonstrating a favorable trade-off between accuracy and efficiency.

\section{Conclusion}
This work presents ViRefSAM, a novel framework that effectively enhances the Segment Anything Model (SAM) for remote sensing segmentation by introducing visual reference guidance. By leveraging several annotated reference images instead of relying on handcrafted prompts, ViRefSAM enables automatic class-specific object segmentation across complex RS scenes, without modifying the original SAM architecture. The proposed Visual Contextual Prompt Encoder extracts object-aware embeddings from reference images to guide the SAM mask decoder, while the Dynamic Target Alignment Adapter further improves task-specific focus by injecting class-specific semantic clues into SAM’s image features. Extensive experiments on three few-shot segmentation benchmarks, iSAID-5$^i$, LoveDA-2$^i$, and COCO-20$^i$, validate the generalization ability and effectiveness of our method. ViRefSAM consistently outperforms existing few-shot segmentation methods, demonstrating its strong potential in addressing the prompt-efficiency and adaptability challenges of SAM in remote sensing applications.




%
\bibliographystyle{IEEEtran}
\bibliography{IEEEabrv,refs}
\vfill

\end{document}